\documentclass[10pt,twocolumn,letterpaper]{article}

\usepackage{cvpr}              % To produce the CAMERA-READY version
\usepackage{graphicx}
\usepackage{amsmath}
\usepackage{amssymb}
\usepackage{booktabs}
\usepackage{makecell}

\usepackage[pagebackref,breaklinks,colorlinks]{hyperref}
\usepackage{bbding}
\usepackage{multirow}

\usepackage[capitalize]{cleveref}
\crefname{section}{Sec.}{Secs.}
\Crefname{section}{Section}{Sections}
\Crefname{table}{Table}{Tables}
\crefname{table}{Tab.}{Tabs.}

\def\cvprPaperID{2611} % *** Enter the CVPR Paper ID here
\def\confName{CVPR}
\def\confYear{2023}

\usepackage{amsmath,amsfonts,bm}

\def\eqref#1{equation~\ref{#1}}
\def\1{\bm{1}}

\def\mC{{\bm{C}}}

\def\mF{{\bm{F}}}

\DeclareMathAlphabet{\mathsfit}{\encodingdefault}{\sfdefault}{m}{sl}
\SetMathAlphabet{\mathsfit}{bold}{\encodingdefault}{\sfdefault}{bx}{n}

\newcommand{\Ls}{\mathcal{L}}

\begin{document}

\title{Inferring and Leveraging Parts from Object Shape for \\
Improving Semantic Image Synthesis}

\author{
\normalsize
Yuxiang Wei$^{1, 2}$ \quad
Zhilong Ji$^{3}$ \quad
Xiaohe Wu$^{1}$\textsuperscript{(\Envelope)} \quad
Jinfeng Bai$^{3}$ \quad
Lei Zhang$^{2}$ \quad
Wangmeng Zuo$^{1}$ \\
\normalsize
$^1$Harbin Institute of Technology \quad $^2$ The Hong Kong Polytechnic University  \quad $^3$Tomorrow Advancing Life  \\
{\tt\small{yuxiang.wei.cs@gmail.com} } \quad {\tt{\small\{wuxiaohe, wmzuo\}@hit.edu.cn}} \quad {\tt{\small cslzhang@comp.polyu.edu.hk }}  \\
}

\twocolumn[{%
\renewcommand\twocolumn[1][]{#1}%
\maketitle
\begin{center}
    \vspace{-2em}
    \centering
    \captionsetup{type=figure}
    \includegraphics[width=0.99 \linewidth]{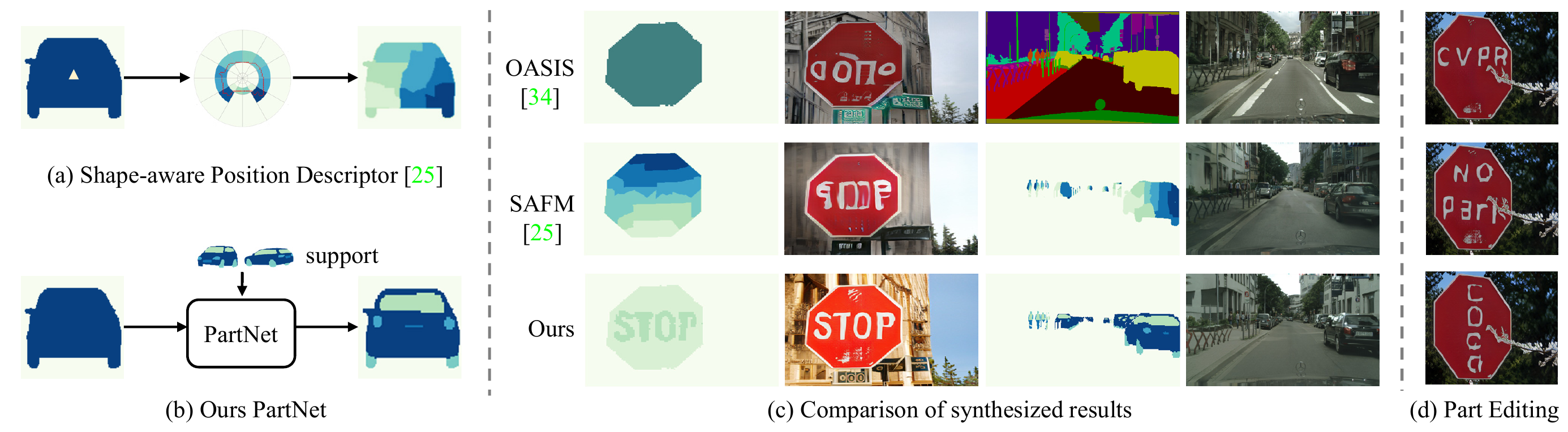}
    \vspace{-1em}
    \captionof{figure}{Illustration of our iPOSE for part map prediction and semantic image synthesis. (a) Calculation of SPD \cite{lv2022semantic}. (b) Our PartNet for part map prediction. (c) Comparison of results with existing methods~\cite{schonfeld2021you, lv2022semantic}. (d) Part editing results of our iPOSE.}
    \label{fig:compare}
\end{center}%
}]

% \maketitle

%%%%%%%%% ABSTRACT
\begin{abstract}

Despite the progress in semantic image synthesis, it remains a challenging problem to generate photo-realistic parts from input semantic map.
Integrating part segmentation map can undoubtedly benefit image synthesis, but is bothersome and inconvenient to be provided by users. 
To improve part synthesis, this paper presents to \textbf{i}nfer \textbf{P}arts from \textbf{O}bject \textbf{S}hap\textbf{E} (iPOSE) and leverage it for improving semantic image synthesis.
However, albeit several part segmentation datasets are available, part annotations are still not provided for many object categories in semantic image synthesis. 
To circumvent it, we resort to few-shot regime to learn a PartNet for predicting the object part map with the guidance of pre-defined support part maps. 
PartNet can be readily generalized to handle a new object category when a small number (e.g., $3$) of support part maps for this category are provided. 
Furthermore, part semantic modulation is presented to incorporate both inferred part map and semantic map for image synthesis.
Experiments show that our iPOSE not only generates objects with rich part details, but also enables to control the image synthesis flexibly.
And our iPOSE performs favorably against the state-of-the-art methods in terms of quantitative and qualitative evaluation. 
Our code will be publicly available at \url{https://github.com/csyxwei/iPOSE}.

\end{abstract}

%%%%%%%%% BODY TEXT
\section{Introduction}
\label{sec:intro}

Semantic image synthesis allows to generate an image with input semantic map, which provides significant flexibility for controllable image synthesis.
Recently, it has attracted intensive attention due to its wide applications, \eg, content generation and image editing~\cite{park2019semantic, shi2022retrieval}, and extensive benefits for many vision tasks~\cite{schonfeld2021you}.

Albeit rapid progress has been made in semantic image synthesis~\cite{isola2017image, qi2018semi,wang2018high,park2019semantic,schonfeld2021you,shi2022retrieval,lv2022semantic,wang2022pretraining,wang2022semantic}, it is still challenging to generate photo-realistic parts from the semantic map. 
Most methods~\cite{park2019semantic, schonfeld2021you} tackle semantic image synthesis with a spatially adaptive normalization architecture, while other frameworks are also explored, such as StyleGAN~\cite{li2021collaging} and diffusion model~\cite{wang2022pretraining, wang2022semantic}, \etc.
However, due to the lack of fine-grained guidance (\eg, object part information), these methods only exploited the object-level semantic information for image synthesis, and usually failed to generate photo-realistic object parts (see the top row of Fig.~\ref{fig:compare}(c)). 
SAFM~\cite{lv2022semantic} adopted the shape-aware position descriptor to exploit the pixel's position feature inside the object.
However, as illustrated in Fig.~\ref{fig:compare}(a), the obtained descriptor tends to be only region-aware instead of part-aware, leading to limited improvement in synthesized results (see the middle row of Fig.~\ref{fig:compare}(c)).

To improve image parts synthesis, one straightforward solution is to ask the user to provide part segmentation map and integrate it into semantic image synthesis during training and inference. 
However, it is bothersome and inconvenient for users to provide it, especially during inference.
Fortunately, with the existing part segmentation datasets~\cite{de2021part}, we propose a method to \textbf{i}nfer \textbf{P}arts from \textbf{O}bject \textbf{S}hap\textbf{E} (iPOSE), and leverage it for improving semantic image synthesis.
Specifically, based on these datasets, we first construct an object part dataset that consists of paired (object shape, object part map) to train a part prediction network (PartNet).
Besides, although the part dataset contains part annotations for several common object categories, many object categories in semantic image synthesis are still not covered.
To address this issue, we introduce the few shot mechanism to our proposed PartNet.
As shown in Fig.~\ref{fig:compare}(b), for each category, a few annotated object part maps are selected as pre-defined supports to guide the part prediction. 
Cross attention block is adopted to aggregate the part information from support part maps.
Benefited from the few shot setting, our PartNet can be readily generalized to handle a new object category not in the object part dataset. 
Particularly, we can manually label $k$ object part maps for this category (\eg, 3)  as supports, and use them to infer the part map without fine-tuning the part prediction model.

By processing each object in the semantic map, we obtain the part map. 
With that, we further present a part semantic modulation (PSM) residual block to incorporate the part map with semantic map to improve the image synthesis.
Specifically, the part map is first used to modulate the normalized activations spatially to inject the part structure information.
Then, to inject the semantic texture information, the semantic map and a randomly sampled 3D noise are further used to modulate the features with the SPADE module~\cite{park2019semantic}.
We find that performing part modulation and semantic modulation sequentially disentangles the structure and texture for image synthesis, and is also beneficial for generating images with realistic parts.
Additionally, to facilitate model training, a global adversarial loss and an object-level CLIP style loss~\cite{zhang2022towards} are further introduced to encourage model to generate photo-realistic images.

Experiments show that our iPOSE can generate more photo-realistic parts from the given semantic map, while having the flexibility to control the generated objects (as illustrated in Fig.~\ref{fig:compare} (c)(d)). 
Both quantitative and qualitative evaluations on different datasets further demonstrate that our method performs favorably against the state-of-the-art methods.

The contributions of this work can be summarized as:
\vspace{-0.6em}
\begin{itemize}
    \setlength{\itemsep}{0pt}
    \setlength{\parsep}{0pt}
    \setlength{\parskip}{0pt}
    \item We propose a method iPOSE to infer parts from object shape and leverage them to improve semantic image synthesis. Particularly, a PartNet is proposed to predict the part map based on a few support part maps, which can be easily generalized to new object categories.
    \item A part semantic modulation Resblock is presented to incorporate the predicted part map and semantic map for image synthesis. And global adversarial and object-level CLIP style losses are further introduced to generate photo-realistic images.
    \item Experimental results show that our iPOSE performs favorably against state-of-the-art methods and can generate more photo-realistic results with rich part details.
\end{itemize}

\section{Related Work}

\subsection{Semantic Image Synthesis}

Semantic image synthesis predicts image from the given semantic map. 
With the development of generative models, many methods have been proposed to solve this problem~\cite{chen2017photographic, qi2018semi, li2019diverse, park2019semantic, schonfeld2021you, wang2018high, ntavelis2020sesame, shi2022retrieval, lv2022semantic, wang2022semantic, wang2022pretraining, tang2020local, wang2021image, zhu2020sean, liu2019learning, isola2017image, li2022dual}.
Pix2pix~\cite{isola2017image} first explored a conditional GAN for translating semantic map to a real image,
Instead of taking semantic map as input directly, SPADE~\cite{park2019semantic} proposed to use it to modulate the features layer-wisely by spatially adaptive denormalization.
CC-FPSE~\cite{liu2019learning} leveraged semantic map to predict the spatial variant convolutional kernels, which were used to generate the intermediate feature maps.
While SC-GAN~\cite{wang2021image} transformed semantic map to semantic vector, and used it for semantic render generation.
RESAIL~\cite{shi2022retrieval} proposed to retrieve and compose a guidance image based on the given semantic map, and incorporated the image to guide the image synthesis.
SAFM~\cite{lv2022semantic} calculated a shape-aware position descriptor for each object in semantic map, and proposed a semantic-shape adaptive feature modulation block to improve object synthesis.

In addition to semantic injection, different networks~\cite{wang2018high, liu2019learning, ntavelis2020sesame, tang2020local, li2021collaging, wang2022semantic} have also been explored in semantic image synthesis, such as multi-scale discriminator~\cite{wang2018high}, feature-pyramid discriminator~\cite{liu2019learning}, and semantic-related discriminator~\cite{ntavelis2020sesame, schonfeld2021you}, \etc.
Besides, LGGAN~\cite{tang2020local} explored the local context information and introduced a local pathway in the generator for details synthesizing.
CollogeGAN~\cite{li2021collaging} used the StyleGAN~\cite{karras2019style,liu2022survey} as the generator to improve visual quality and also explored the local context with class-specific models. 
Recently, SDM~\cite{wang2022semantic} proposed a semantic diffusion model to generate images with an iterative denoising process conditioned on the semantic maps. 

Most methods have only exploited the object-level semantic information for image synthesis, resulting in unrealistic parts synthesis. 
Albeit SAFM~\cite{lv2022semantic} exploited to use the object's pixel position feature for image synthesis, the obtained descriptor is only region-aware and also limited.
In contrast, our iPOSE infers the part map from the given semantic map and leverage it to improve parts synthesis.
It not only generates objects with realistic parts, but also enables to control the image synthesis flexibly.

\subsection{Few Shot Segmentation}

Few shot segmentation~\cite{tian2020prior,tian2022generalized,shaban2017one,zhang2020sg,zhang2019canet,li2021adaptive,xie2021scale,wu2021learning,wang2019panet,zhang2019pyramid,shi2022dense,ao2022few,liu2022learning} aims to predict a dense segmentation for new class with only a few annotated support images.
Shaban \etal~\cite{shaban2017one} introduced the few shot segmentation task and proposed a two-branch networks to generate classifier weights from the support images for query image segmentation. 
Rakelly \etal~\cite{rakelly2018conditional}  constructed the global conditioning prototype from the support set and concatenated it to the query representation.
Following this prototype paradigm,  
MM-Net~\cite{wu2021learning} introduced a set of learnable memory embeddings to store the meta-class information during training and transfer them to novel classes during the inference stage.
Besides, several methods have also exploited pixel-level information for few shot segmentation.
PGNet~\cite{zhang2019pyramid} used a graph attention unit to build pixel level dense similarity between the query and support images.
PFENet~\cite{tian2020prior} calculated the cosine similarity on high-level features without trainable parameters to create a prior mask and introduced a feature enrichment module to reduce the spatial inconsistency between the query and support samples.
DCAMA~\cite{shi2022dense} further proposed a cross attention weighted mask aggregation with multi-scale mechanism for few shot segmentation.

\subsection{Part Segmentation}

Part segmentation has also received considerable attention in semantic segmentation. 
Earlier methods usually treated part segmentation as a semantic segmentation problem, and most researches focused on part segmentation for humans~\cite{liu2018cross, li2017holistic, ruan2019devil, gong2018instance, li2018multi}.
For example, PGN~\cite{gong2018instance} proposed a part grouping network to solve multi-person human parsing in a unified network, which contains two twinned grouping tasks, \ie, semantic part segmentation and instance aware edge detection. 
CE2P~\cite{ruan2019devil} introduced a simple yet efficient context embedding with edge perceiving framework by leveraging the useful properties to conduct human parsing.
Besides human, several other categories have also been studied, such as faces~\cite{lin2019face} and animals~\cite{wang2015joint, chen2014detect}.
Recently, PPS~\cite{de2021part} proposed two part annotated datasets, \ie, Cityscapes PPS and Pascal VOC PPS for panoptic part segmentation. 
Subsequently, Panoptic-PartFormer~\cite{li2022panoptic} proposed a unify model to learn thing, stuff, and part prediction tasks simultaneously in an end-to-end manner.

\begin{figure}[t]
\centering
\includegraphics[width=1\linewidth]{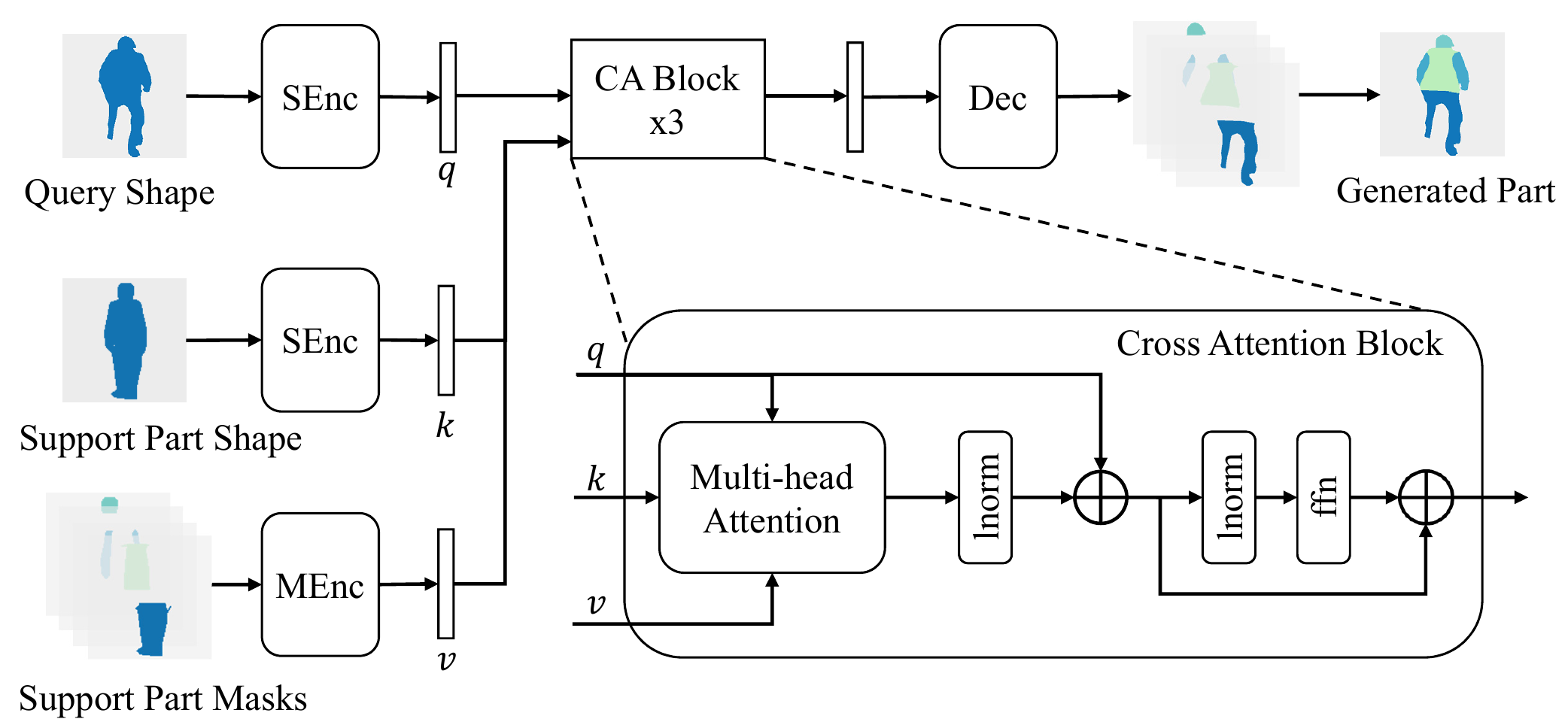}
\vspace{-1.5em}
\caption{Illustration of our PartNet. The support part map is first decomposed into the support part shape and the support part masks as inputs. Cross attention block is adopted to aggregate the part information from the support features.}
\vspace{-1.5em}
\label{fig:partnet}
\end{figure}

\begin{figure*}[t]
\centering
\includegraphics[width=1\linewidth]{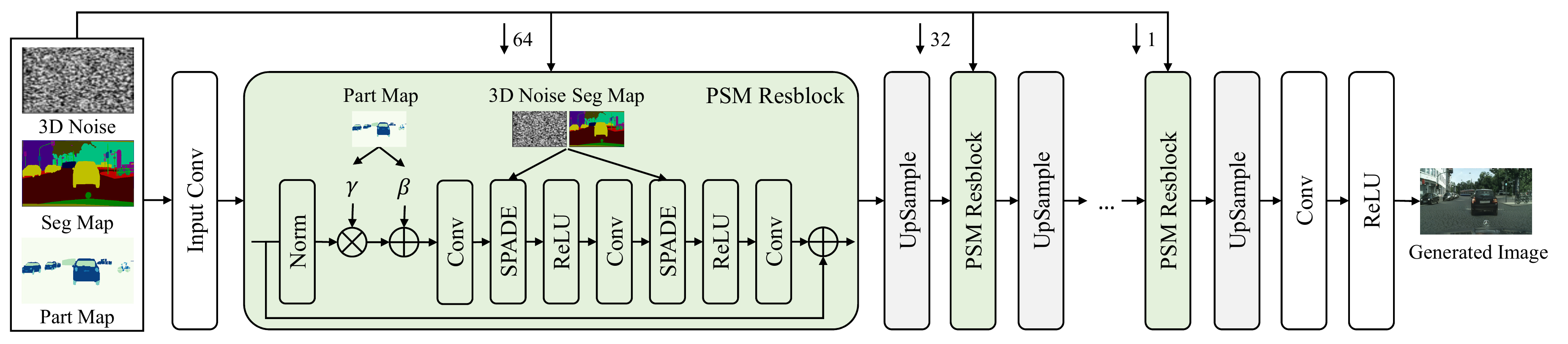}
\vspace{-1.5em}
\caption{Architecture of our generator. It takes part map, semantic map and 3D noise as the input, while performing part semantic modulation for image synthesis.}
\label{fig:generator}
\vspace{-1.5em}
\end{figure*}

\section{Proposed Method}

Given a semantic map $M \in \{0,1\}^{H \times W \times C}$ with $C$ categories, semantic image synthesis aims to generate the corresponding images $\hat{I} \in \mathbb{R}^{H \times W \times 3}$. 
One main challenge is to generate photo-realistic parts from the semantic map.
To address this problem, we present to \textbf{i}nfer \textbf{P}arts from \textbf{O}bject \textbf{S}hap\textbf{E} (iPOSE), and leverage it for improving semantic image synthesis. 
Specifically, a PartNet is first proposed to infer the part map from the object shapes with the guidance of pre-defined support part maps (Sec.~\ref{sec:part}).
Following~\cite{wang2018high, lv2022semantic}, we adopt the instance-level segmentation map to obtain the shape of each object in $M$.
With the part map $P$, we further present a part semantic modulation ResBlock to incorporate it with the semantic map $M$ and the 3D noise $Z$ to modulate the generation process (Sec.~\ref{sec:syn}). 
To facilitate model training, several loss terms are introduced to encourage the model to generate photo-realistic images (Sec.~\ref{sec:loss}).

\subsection{Inferring Parts from Object Shape}
\label{sec:part}

To improve the part synthesis, 
we first present to learn a PartNet based on existing part segmentation datasets~\cite{de2021part}, which infers the part map $P$ from the given semantic map $M$.
Specifically, the semantic map $M$ can be decomposed into several object shapes $\{(O_i, y_i)\}$, where $O_i$ denotes the cropped object shape and $y_i$ is the corresponding category.
For each object shape $O_i$, PartNet predicts the corresponding object part map $P_i$.
To handle those objects with novel categories (\ie, category not in part training datasets~\cite{de2021part}), we further introduce the few shot mechanism to learn the PartNet.
And for each category $y$, a few object part maps $S_y = \{ S_{y, 0}, \cdots, S_{y, k-1}\}$ are selected as support set to guide the part prediction. 
For brevity, we take $k = 1$ as example, and our design can be easily extended to few shot, \eg, $k=3$ in our implementation.
The architecture of our PartNet is illustrated in Fig.~\ref{fig:partnet}.
For each query object $O_q$, the corresponding support $S_{y_q}$ is decomposed into the support part shape $O^S_{y_q}$ and the support part masks $O^M_{y_q}$ as inputs.
Each mask denotes an object part, \eg, human's head, arm and leg, \etc.

\noindent \textbf{Architecture of PartNet.} 
Following few shot segmentation, our PartNet aims to predict each part of $O_q$ based on its similarity with $O^S_{y_q}$ and the part prior $O^M_{y_q}$.
However, there is a lack of texture information for the object shape, which is unsuitable for the pre-trained image encoder.
Instead, we utilize a shape encoder to extract the shape and position information from the object shapes $O_q$ and $O^S_{y_q}$.
A position embedding $\mathbf{p}$~\cite{vaswani2017attention} is further concatenated with $O_q$ and $O^{S}_{y_q}$ as input.
% Position embedding~\cite{vaswani2017attention} is also incorporated with each shape.
%
Besides, to exploit the part information, a mask encoder is further introduced to encode $O^M_{y_q}$,
\vspace{-0.5em}
\begin{equation}
\small
\mathbf{q} = \text{SEnc}(O_q), \ \mathbf{k} = \text{SEnc}(O^S_{y_q}), \ \mathbf{v} = \text{MEnc}(O^M_{y_q}),
\end{equation}
where SEnc and MEnc denote the shape and mask encoder, respectively. 
Additionally, to perceive pixels' relative position of the whole object shape, the multi-scale mechanism is also adopted by the encoders.
Features from different scales are upsampled and concatenated as encoder output.
Furthermore, to aggregate the part information for part prediction, we introduce a cross attention (CA) block.
In particular, a shape position similarity matrix $\mathbf{q}\mathbf{k}^T$ is calculated between $\mathbf{q}$ and $\mathbf{k}$.
The same part region in two object shapes tends to have a large similarity (\eg, head part in $O_q$ and $O^S_{y_q}$).
Then, it multiplies with $\mathbf{v}$ to aggregate the part information.
Layer norm and FFN are also adopted in CA block to improve the performance. 
Finally, the obtained feature is passed through the decoder to generate object part map $P_q$.

\noindent \textbf{Generalizing to novel categories.} 
We pre-train the PartNet on the basis categories of constructed part object dataset (see Sec.~\ref{sec:exp}) and fix it during the following training.
For those objects in $M$ with novel categories (not in basis categories, \eg, stop sign in COCO), our PartNet can be easily generalized to handle these objects.
% %
% %
For example, to predict the part map for stop sign category, we just need to select $k$ stop sign shapes from the COCO dataset, and manually annotate their part maps as supports (\eg, the word STOP and the background).
% %
Then, given a new stop sign shape, our PartNet can leverage the supports to infer the part map without fine-tuning (as shown in Fig.~\ref{fig:compare}).
More details about the pre-training and novel categories can be found in \emph{Suppl}.

\subsection{Leveraging Parts for Semantic Image Synthesis}
\label{sec:syn}

With the part map $P$, we then take it with the semantic map $M$ to generate the image.
Following~\cite{schonfeld2021you}, a random sampled 3D noise $Z$ is also incorporated.
Intuitively, the part map represents the structure information of the generated image, yet the semantic map and 3D noise provide the semantic texture information.
To disentangle the structure with texture for image synthesis, we further present a part semantic modulation Resblock to modulate the activations.

\noindent \textbf{Part Semantic Modulation Resblock.} 
As illustrated in Fig.~\ref{fig:generator}, to achieve a disentangled synthesis, we inject the part and semantic sequentially.
Firstly, to inject the structure information, the part map $P$ is used to modulate the activations.
To encourage it to guide the structure synthesis independently, the modulation is performed spatially, \ie, $\gamma \in \mathbb{R}^{H \times W \times 1}$ and $\beta \in \mathbb{R}^{H \times W \times 1}$.
Besides, for those regions in the part map without structure information, we introduce additional noise to increase the diversity.
Then, SPADE~\cite{park2019semantic} is adopted to take semantic map $M$ and 3D noise $Z$ together to modulate the activations to inject the semantic texture information.
By the separate modulation, our iPOSE can disentangle structure and texture successfully, and we also found that performing part modulation and semantic modulation sequentially is beneficial for generating images with photo-realistic parts.

\noindent \textbf{Generator.} 
Following \cite{schonfeld2021you}, we stack the part semantic modulation (PSM) Resblocks and upsampling layers to constitute our generator (as shown in Fig.~\ref{fig:generator}).
The semantic map $M$, part map $P$ and 3D noise $Z$ are resized and fed to each PSM Resblock to guide the image synthesis.
\begin{equation}
\small
\hat{I} = G(M, P, Z).
\label{eq: synthesis}
\end{equation}
\vspace{-2em}
%
% where $P$ denotes the part map inferred from semantic map.

\begin{figure*}[t]
\centering
\includegraphics[width=1\linewidth]{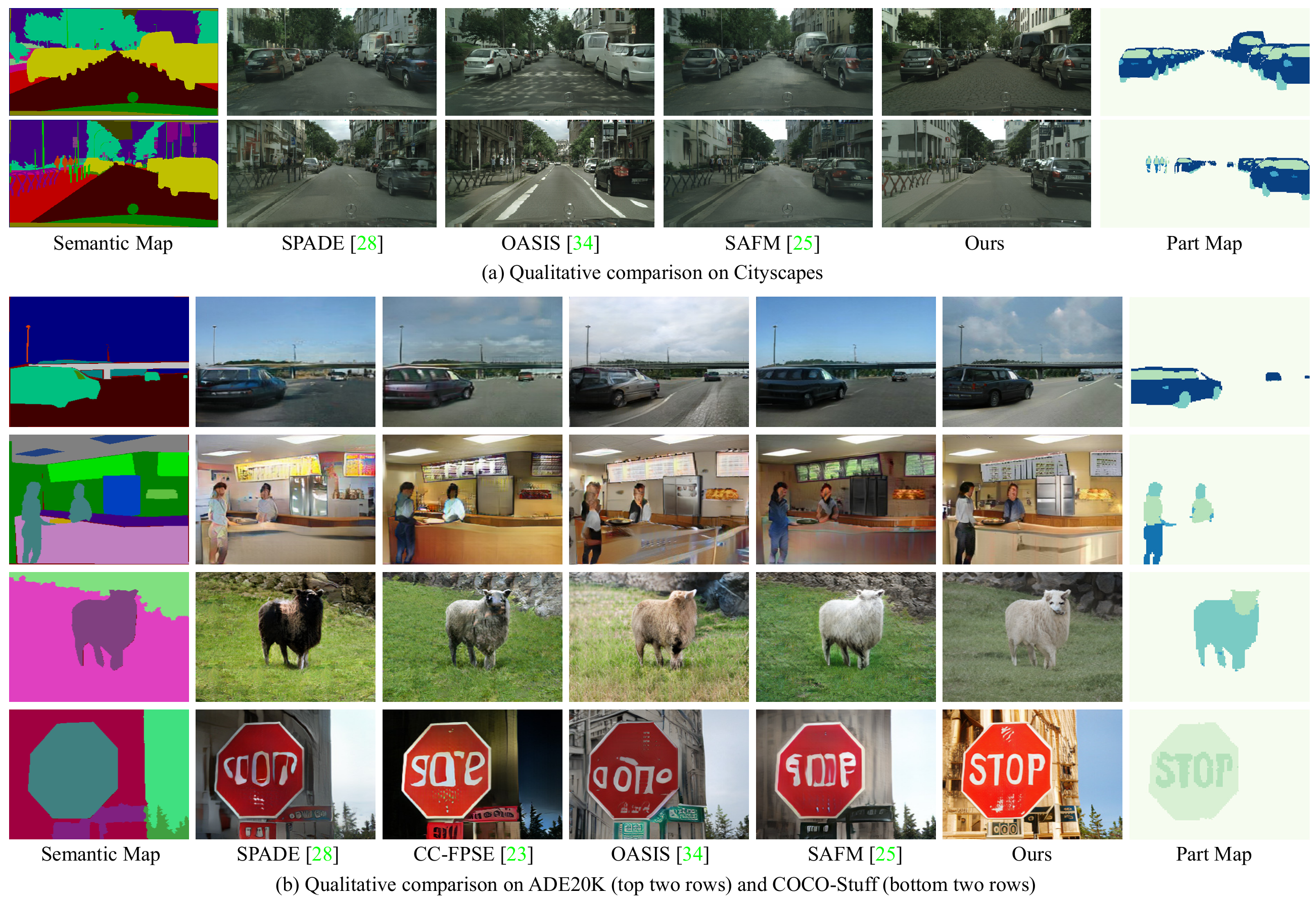}
\vspace{-2em}
\caption{Visual comparisons on the Cityscapes (1st $\sim$ 2nd rows), ADE20K (3rd $\sim$ 4th row) and COCO-stuff (5th $\sim$ 6th rows) datasets.}
\vspace{-1.5 em}
\label{fig:qualitative_comparison}
\end{figure*}

\subsection{Learning Objective}
\label{sec:loss}

We incorporate several loss terms to encourage the model to generate photo-realistic images.
Following OASIS~\cite{schonfeld2021you}, we first introduce the $(N+1)$-class adversarial loss $\mathcal{L}^{N+1}_{G}$, $\mathcal{L}^{N+1}_{D}$, and the LabelMix loss $\mathcal{L}_{label}$ to train our model. 
In addition, to improve the quality of synthesized objects, we suggest a global adversarial loss~\cite{park2019semantic} and an object-based CLIP style loss~\cite{zhang2022towards}.

\noindent \textbf{Global Adversarial Loss.} 
Although the $(N+1)$-class adversarial loss can improve the semantic alignment of each pixel, it lacks the global constraint on images.
To further improve the generator to synthesize realistic images, we suggest a global adversarial loss, 
\begin{equation}
\small
\mathcal{L}^{global}_{D} \!=\! \mathbb{E}_I[\log D(I)] + 
\mathbb{E}_{M, Z}[\log (1- D(G(M, P, Z))], 
\end{equation}
\begin{equation}
\small
\mathcal{L}^{global}_{G} =  \mathbb{E}_{M, Z}[\log (1- D(G(M, P, Z))],
\end{equation}
where $Z$ is the 3D noise, and $I$, $M$ denote the real image and its corresponding semantic map, respectively. 

\noindent \textbf{Object-level CLIP Style Loss.}
To facilitate the learning of object synthesis, we also introduce an object-level CLIP style loss~\cite{zhang2022towards}.
Specifically, CLIP image encoder is adopted as the feature extractor, and extracts the intermediate tokens of images ($I$ and $\hat{I}$) from the $l$-th layer.
Then we align each token of generated image $\mF_{\hat{I}}$ with the closest token of real image $\mF_{I}$, where $\mF_{\hat{I}}=\{\mF_{\hat{I}}^1,\dots,\mF_{\hat{I}}^n\}$ and $\mF_{I}=\{\mF_{I}^1,\dots,\mF_{I}^n\}$ are the extracted tokens,
\begin{equation}
\label{l_style}
\small
\Ls_{style} = \max \Big(\frac 1 n \sum_i \min_j \mC_{i,j}, \frac 1 m \sum_j \min_i \mC_{i,j} \Big),
\end{equation}
where $\mC$ is the cost matrix to measure the token-wise distances from $F_I$ to $F_{\hat{I}}$, and each element of $\mC$ is given by,
\begin{align}
\small
\label{eq_cost}
    \mC_{i,j} = 1 - \frac {\mF_{\hat{I}}^i \cdot \mF_{I}^j} {|\mF_{\hat{I}}^i| |\mF_{I}^j|}.
\end{align}
It is worthy noting that, to emphasize object synthesis, we calculate the $\mathcal{L}_{style}$ with the objects cropped from $I$ and $\hat{I}$.

The overall learning objective can be summarized as,
\begin{equation}
\small
\mathcal{L}_{G} =  \mathcal{L}^{N+1}_{G} + \lambda_{global} \mathcal{L}^{global}_{G} + \lambda_{style} \mathcal{L}_{style},
\end{equation}
\begin{equation}
\small
\mathcal{L}_{D} = \mathcal{L}^{N+1}_{D} + \lambda_{global} \mathcal{L}^{global}_{D} + \lambda_{label} \mathcal{L}_{label},
\end{equation}
where $\lambda_{global}$, $\lambda_{style}$ and $\lambda_{label}$ denote the trade-off parameters for different loss terms.

\begin{figure*}[t]
\centering
\includegraphics[width=1\linewidth]{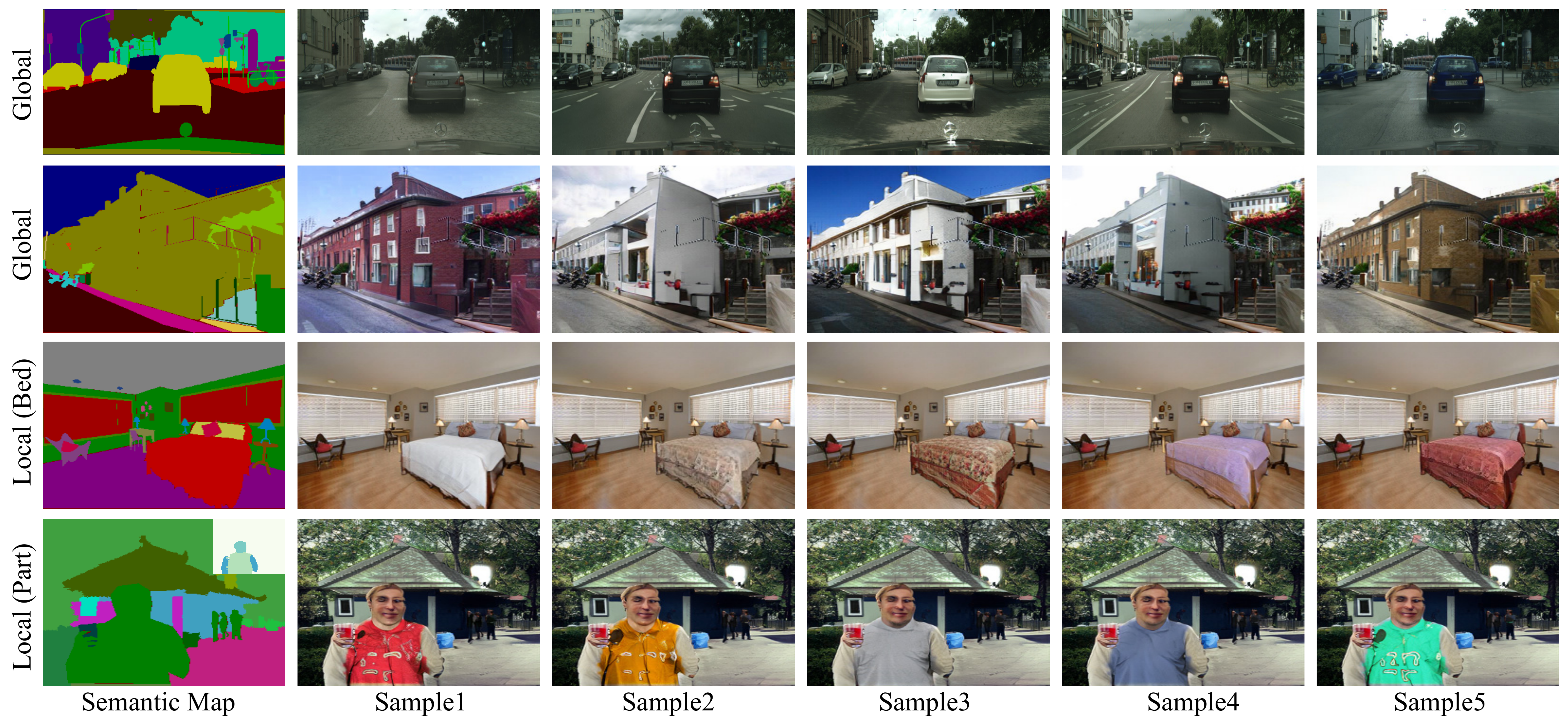}
\vspace{-2em}
\caption{Multi-modal synthesis results of our iPOSE. By injecting different 3D noises into the assigned region, our iPOSE can achieve global-level (first 2 rows), object-level (3rd row), and also part-level (last row) image editing.}
\vspace{-0.5em}
\label{fig:style_diversity}
\end{figure*}

\begin{figure}[t]
\centering
\includegraphics[width=1\linewidth]{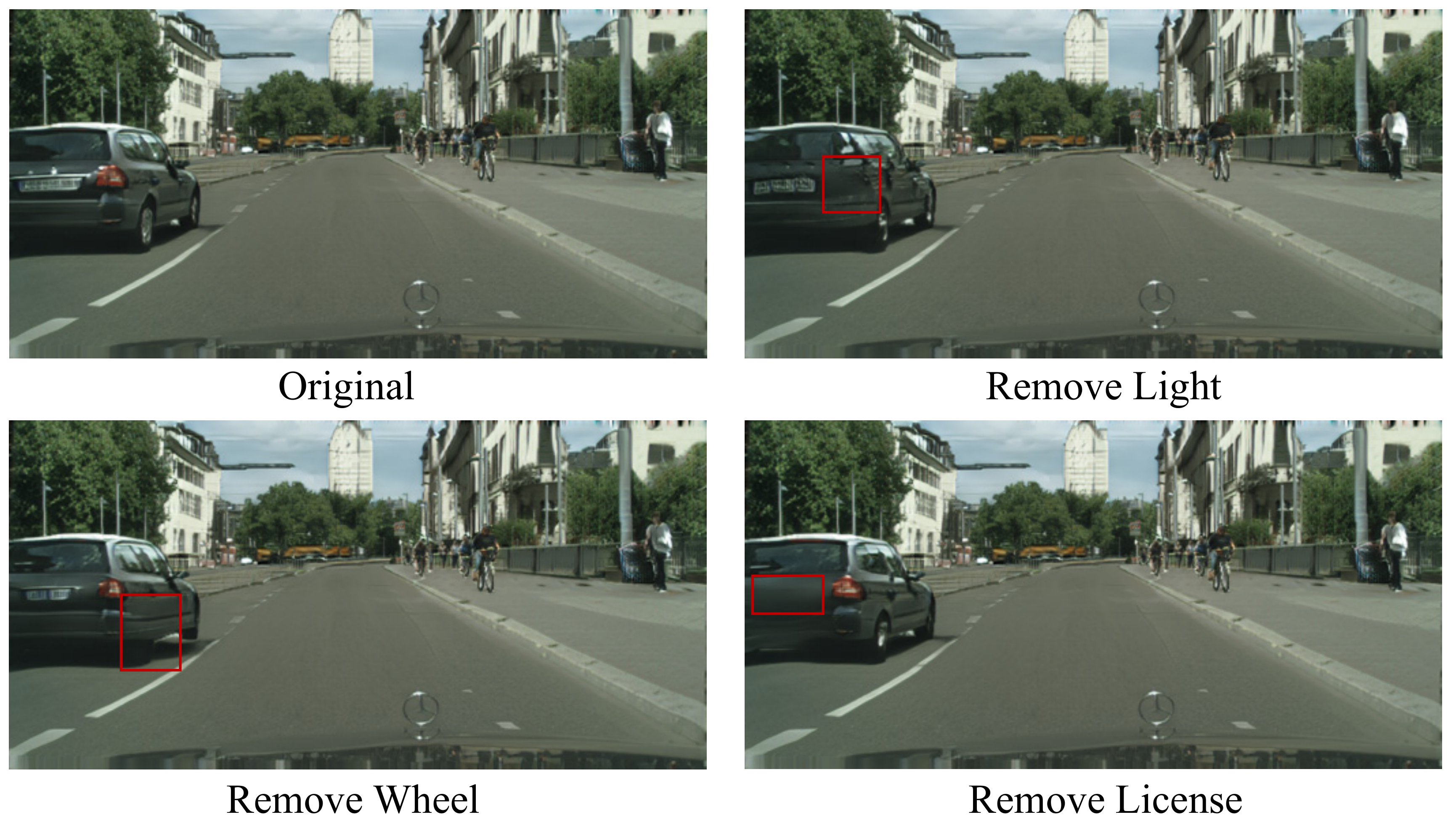}
\vspace{-1.5em}
\caption{Part editing results of our iPOSE. Our method allows editing parts of generated objects through the support part maps. For example, we can remove the light, wheel, or license of the car.}
\vspace{-2em}
\label{fig:part_diversity}
\end{figure}

\begin{table*}[t]
{\small
    \begin{center}
    \caption{Quantitative comparison with existing methods on different datasets. $\uparrow$ ($\downarrow$) denotes the higher (lower) is better.}
    \vspace{-1em}
    \label{tab:results}
    \resizebox{0.99\linewidth}{!}{
        \setlength{\tabcolsep}{1.2em}
        \renewcommand{\arraystretch}{1.0}
        \begin{tabular}{lccccccccc}
            \toprule
            \multirow{2}{*}{Method} &
            \multicolumn{3}{c}{Cityscapes} &
            \multicolumn{3}{c}{ADE20K} &
            \multicolumn{3}{c}{COCO-Stuff} \\
            \cmidrule(llr){2-4}
            \cmidrule(llr){5-7}
            \cmidrule(llr){8-10}
            & FID ($ \downarrow $) & AC ($ \uparrow $) & mIOU ($ \uparrow $) & FID ($ \downarrow $) & AC ($ \uparrow $) & mIOU ($ \uparrow $) & FID ($ \downarrow $) & AC ($ \uparrow $) & mIOU ($ \uparrow $) \\
            \midrule
            
            SIMS~\cite{qi2018semi} & 49.7 & 75.5 & 47.2 & n/a & n/a & n/a  & n/a & n/a & n/a \\
            
            SPADE~\cite{park2019semantic}  & 71.8 & 81.9 & 62.3  & 33.9 & 79.9 & 38.5 &  22.6 & 67.9 & 37.4 \\
            
            CC-FPSE~\cite{liu2019learning} & 54.3 & 82.3 & 65.5 & 31.7 & 82.9 & 43.7 &  19.2 & 70.7 & 41.6 \\
            
            SC-GAN~\cite{wang2021image}  & 49.5 & 82.5 & 66.9 & 29.3 & 83.8 & 45.2 &  18.1 & 72.0 & 42.0 \\
            
            OASIS~\cite{schonfeld2021you} & 47.7 & n/a & 69.3 & 28.3 & n/a & 48.8 &  \underline{17.0} & n/a & 44.1 \\
            
            RESAIL~\cite{shi2022retrieval} & 45.5 & \textbf{83.2} &  69.7 & 30.2 & 84.8 & 49.3 &  18.3 & 73.1 & \underline{44.7} \\

            SAFM~\cite{lv2022semantic} & 49.5 & \underline{83.1} & 70.4 & 32.8 & \underline{86.6} & \underline{50.1} &  24.6 & \underline{73.4} & 43.3 \\

            SDM~\cite{wang2022semantic} & \underline{42.1} & n/a & \textbf{77.5} & \underline{27.5} & n/a & 39.2 &  n/a & n/a & n/a \\

            Ours & \textbf{41.3} & 82.2 & \underline{70.6} & \textbf{26.9} & \textbf{87.1} & \textbf{53.8} &  \textbf{15.7} & \textbf{74.8} & \textbf{45.1} \\
            
            \bottomrule
            \vspace{0.1pt}
        \end{tabular}
    }
    \end{center}
}
\vspace{-3em}
\end{table*}

\section{Experiments}

\subsection{Experimental settings}
\label{sec:exp}
\noindent \textbf{Dataset for Part Seg.} 
We first construct an object part dataset based on the Cityscapes PPS and Pascal VOC PPS datasets~\cite{de2021part}. 
Specifically, each object part is cropped and resized to 64$\times$64 based on its bounding box, and formed as paired (object shape, object part map). 
There is a total of 21 categories in the dataset.
Following~\cite{shaban2017one}, we use 20 basis categories for PartNet training (\eg, human, car, bus, and sheep, \etc) and use the remaining category as novel category for openset testing (\ie, cat).
To obtain the support object part maps for each category, we use k-means to cluster the training shapes  into $k$ clusters based on the shape similarity metric~\cite{shi2022retrieval}.
Besides, for those novel categories in image synthesis datasets,  we have annotated $k$ support part maps manually to perform part prediction, including the washing machine, van, zebra, cat, and stop sign.
More details can be found in \emph{Suppl}.

\noindent \textbf{Datasets for Semantic Image Synthesis.} 
Following \cite{lv2022semantic, schonfeld2021you, park2019semantic}, we conduct the experiments on Cityscapes~\cite{cordts2016cityscapes}, ADE20K~\cite{zhou2017scene}, and COCO-Stuff~\cite{caesar2018coco}. 
Cityscapes includes 35 semantic categories, and consists of 3,000 training images and 500 validation images. 
ADE20K contains over 20,000 training images and 2,000 validation images with 150 semantic classes. 
COCO-Stuff consists of 118,000 training images and 5,000 validation images.
In our experiments, the images in ADE20K and COCO-Stuff are resized and cropped to 256 $\times$ 256, while those in Cityscapes are processed to 256 $\times$ 512.

\noindent \textbf{Evaluation Metrics.} 
To evaluate our method, we adopt Pixel ACcuracy (AC), mean Intersection-Over-Union (mIOU), and Frechet Inception Distance (FID)~\cite{heusel2017gans} as the metrics. 
AC and mIOU measure the semantic consistency between the generated image and given input~\cite{park2019semantic, schonfeld2021you, lv2022semantic}, and pretrained segmentation models are adopted to compute segmentation accuracy~\cite{chen2017deeplab, xiao2018unified, yu2017dilated}. 
%h
Furthermore, FID evaluates the quality and diversity of generated images.

\noindent \textbf{Implementation Details.} 
We implement our model with Pytorch~\cite{paszke2019pytorch} and train it with 4 Tesla V100 GPUs. 
For PartNet, we set the number of support part maps to be $k = 3$.
For synthesis model, 
Adam~\cite{kingma2014adam} is adopted with $\beta_1$ = 0 and $\beta_2$ = 0.999 and the learning rates are set to 0.0001 for generator and 0.0004 for discriminator. 
The trade-off parameters $\lambda_{global}$, $\lambda_{style}$ and $\lambda_{label}$ are set to 1, 10 and 10, respectively. 
Following~\cite{park2019semantic}, we apply the spectral normalization~\cite{miyato2018spectral} to each layer in both generator and discriminator. 

\begin{figure}[t]
\centering
\includegraphics[width=1\linewidth]{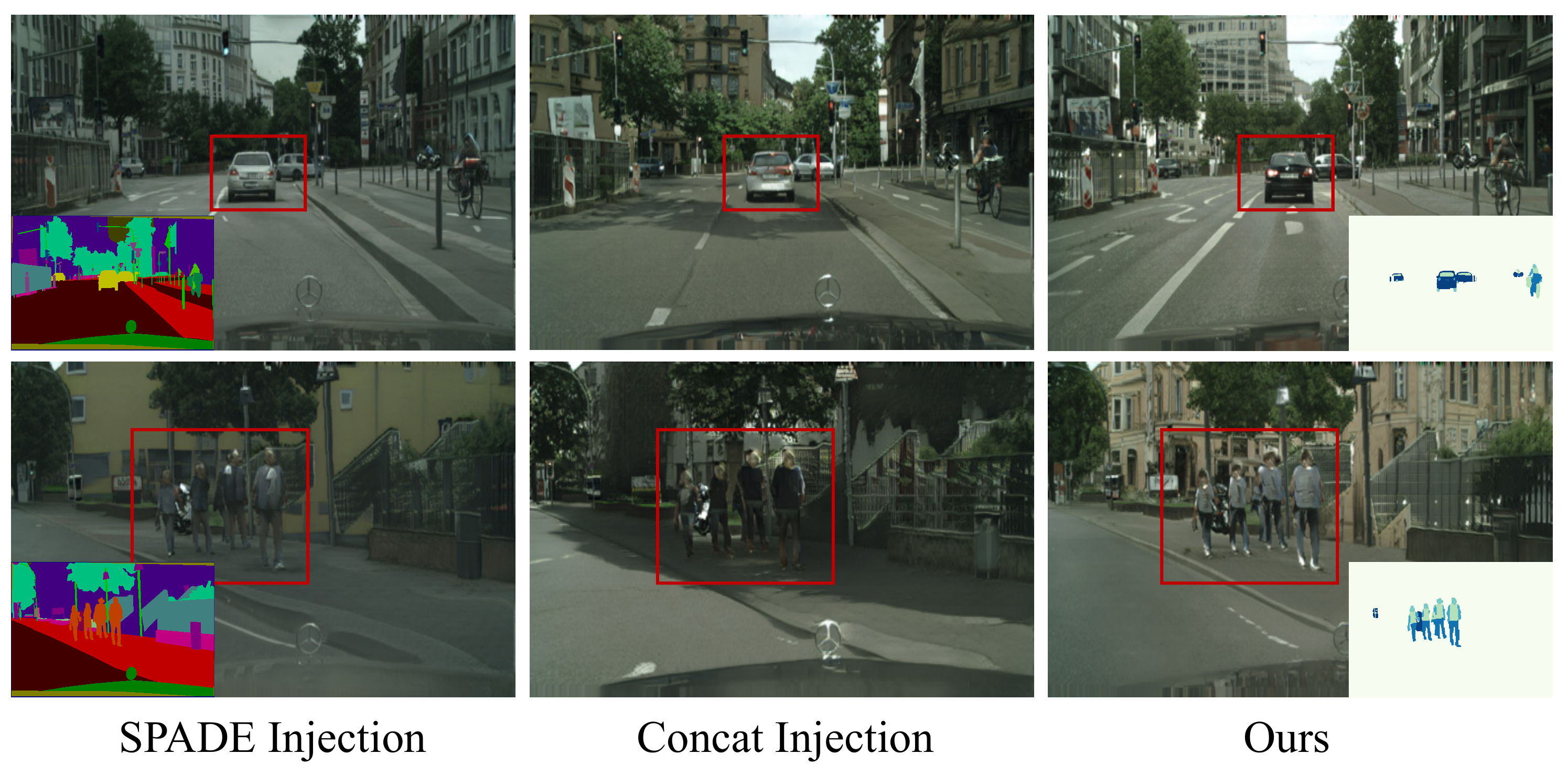}
\vspace{-2em}
\caption{Visual comparisons of different part injection methods. }
\label{fig:ablation_injection}
\vspace{-1.8em}
\end{figure}

\subsection{Qualitative Results}

Fig.~\ref{fig:qualitative_comparison} gives the qualitative comparisons with the SPADE~\cite{park2019semantic}, CC-PFSE~\cite{liu2019learning}, OASIS~\cite{schonfeld2021you}, and SAFM~\cite{lv2022semantic} on the three datasets.
For illustration, the inferred part maps are also shown on the right.
From the figure we can see that, our iPOSE predicts the plausible part map based on the given semantic map.
For the novel categories not in object part training set (\eg, stop sign), our PartNet still works by providing 3 annotated stop sign part maps as supports, demonstrating its generalization ability.
Furthermore, our iPOSE generates images that are semantically consistent with the part maps, and also with high quality and fine details (\eg, light of car, head of sheep, and stop sign).
In contrast, only exploiting the object information, SPADE, CC-FPSE, and OASIS generate images with noticeable artifacts (\eg, car, sheep and stop sign).
Although SAFM explores the shape-aware position descriptor, it is region-aware and lacks part prior, resulting in limited improvement (\eg sheep and stop sign).
In comparison with the competing methods, our iPOSE can generate more photo-realistic images, clearly demonstrating its superiority.

In addition, our iPOSE learns to disentangle the structure and texture for image synthesis, and allows to edit the texture and structure independently.
Fig.~\ref{fig:style_diversity} illustrates the multi-modal synthesis results by injecting different sampled noises into assigned region. 
As shown in the figure, benefited from the predicted part map, our iPOSE can achieve global-level, object-level, and part-level image editing.
Moreover, we can edit the structure of generated objects.
As shown in Fig.~\ref{fig:part_diversity}, we can remove the license, light, and wheel of generated images, thus enabling the users to control the image synthesis more flexibly.
Similar results in Fig.~\ref{fig:compare}(d).
More qualitative results are shown in the \emph{Suppl}.

\begin{table}[t]
\begin{center}
\caption{User study on different datasets. The numbers indicate the percentage (\%) of volunteers who favor the results of our method over those of the competing methods. }
\vspace{-0.8em}
\label{tab:user_study}
\resizebox{1\linewidth}{!}{
    \setlength{\tabcolsep}{1.2em}
    \begin{tabular}{lcccc}
        \toprule
        Dataset & \makecell[c]{Ours vs. \\ SPADE} & \makecell[c]{Ours vs. \\ CC-FPSE}  & \makecell[c]{Ours vs. \\ OASIS} & \makecell[c]{Ours vs. \\ SAFM} \\
        \midrule
        Cityscapes~\cite{cordts2016cityscapes} & 75.6 & 64.8 & 67.3 & 58.6 \\
        ADE20K~\cite{zhou2017scene} & 72.3 & 62.4 & 60.5 & 57.6 \\
        COCO-Stuff~\cite{caesar2018coco} & 67.7 & 57.2 & 57.4 & 61.9 \\
        \bottomrule
    \end{tabular}
}
\vspace{-1.5em}
\end{center}
\end{table}

\begin{table}[t]
\begin{center}
\caption{Ablation studies on the losses, part map, and different part injection methods. With the introduced $\mathcal{L}^g_{G/D}$, $\mathcal{L}_{style}$ and the proposed part semantic modulation, our method achieves better quantitative performance.}
\vspace{-0.8em}
\resizebox{1.0\linewidth}{!}{
    \label{tab:ablations_results}
    \setlength{\tabcolsep}{2.5mm}
    \begin{tabular}{ccccccc}
        \toprule
        Part Inject & $\mathcal{L}^g_{G/D}$ & $\mathcal{L}_{style}$ & FID$\left(\downarrow\right)$ & mIOU$\left(\uparrow\right)$ & AC$\left(\uparrow\right)$ & obj FID $\left(\downarrow\right)$  \\
        \midrule
         &  &  & 47.7 & 66.9 & 81.5 & 44.1 \\
         & \checkmark &  & 43.6 & 66.7 & 81.9 & 39.2 \\
         & \checkmark & \checkmark & 42.8 & 70.5 & 82.1 & 37.5 \\
        Ours PSM & \checkmark & \checkmark & \textbf{41.3} & \textbf{70.6} & \textbf{82.2} & \textbf{30.4} \\
        \midrule
        SPADE & \checkmark & \checkmark & 42.9 & 69.9 & 81.9 & 31.2 \\
        Concat & \checkmark & \checkmark & 42.7 & 70.6 & 82.0 & 31.5 \\
        \bottomrule
    \end{tabular}
}	
\vspace{-2em}
\end{center}
\end{table}

\subsection{Quantitative Results}

Table~\ref{tab:results} lists the quantitative comparison between our iPOSE with the state-of-the-art methods~\cite{qi2018semi,park2019semantic,liu2019learning,wang2021image,schonfeld2021you,shi2022retrieval,lv2022semantic,wang2022semantic}.
From the table, our method performs favorably against the competing methods on most datasets in terms of the three metrics.
Benefited from the introduced part map and losses, our iPOSE can generate images with higher quality and finer details, especially for object regions, resulting in a notable improvement of FID metric ($+0.8$, $+0.6$, $+1.3$ on three datasets, respectively). 
%
% Besides, our method achieve better semantic alignment with the input semantic map on ADE20K and COCO-stuff, demonstrating its effectiveness.
%
Besides, the images generated by our iPOSE are better semantically aligned with the input semantic map on ADE20K and COCO-stuff, further demonstrating its effectiveness.

\noindent \textbf{User Study.} 
Following~\cite{park2019semantic}, we conduct the user study to compare with four competing methods~\cite{park2019semantic,liu2019learning,schonfeld2021you,lv2022semantic}.
Specifically, the participants\footnote{The identities will not be recorded.} were given an input segmentation mask and generated results from two different methods, and required to select the image that has better performance in semantic alignment and photo-realistic appearance. 
For each comparison, we randomly generate 500 questions for each dataset, and each question is answered by 10 different workers.
From Table~\ref{tab:user_study}, users tend to favor our results on all the datasets, especially on Cityscapes.

\begin{figure*}[t]
\centering
\includegraphics[width=1\linewidth]{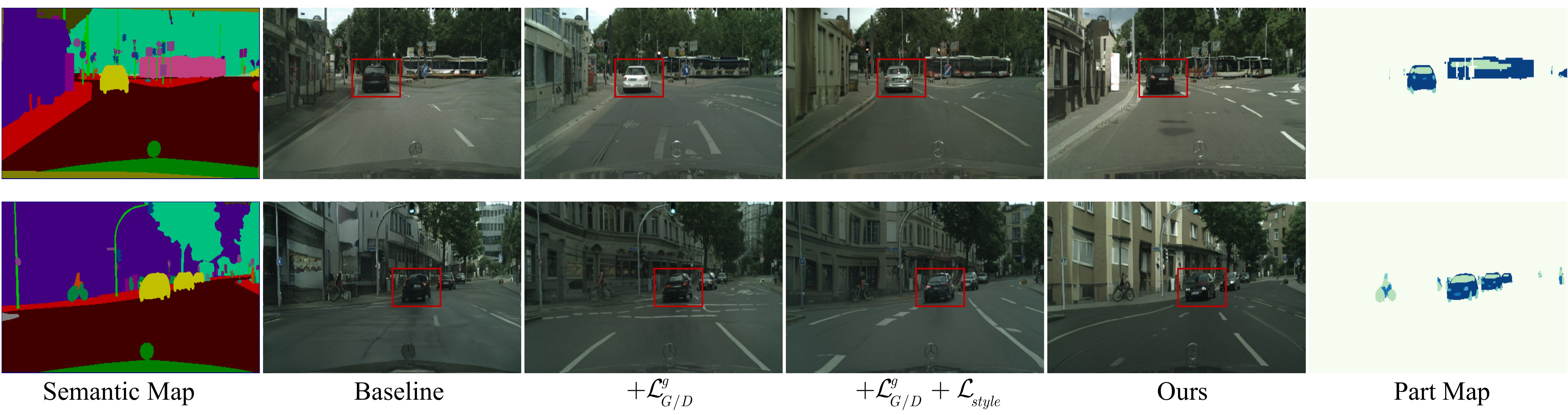}
\vspace{-1.8em}
\caption{Visual comparisons of different variants. Our full model with the introduced losses and part semantic modulation achieves better visual quality, especially for object details.}
\label{fig:ablation_loss}
\vspace{-1em}
\end{figure*}

\begin{figure}[t]
\centering
\includegraphics[width=1\linewidth]{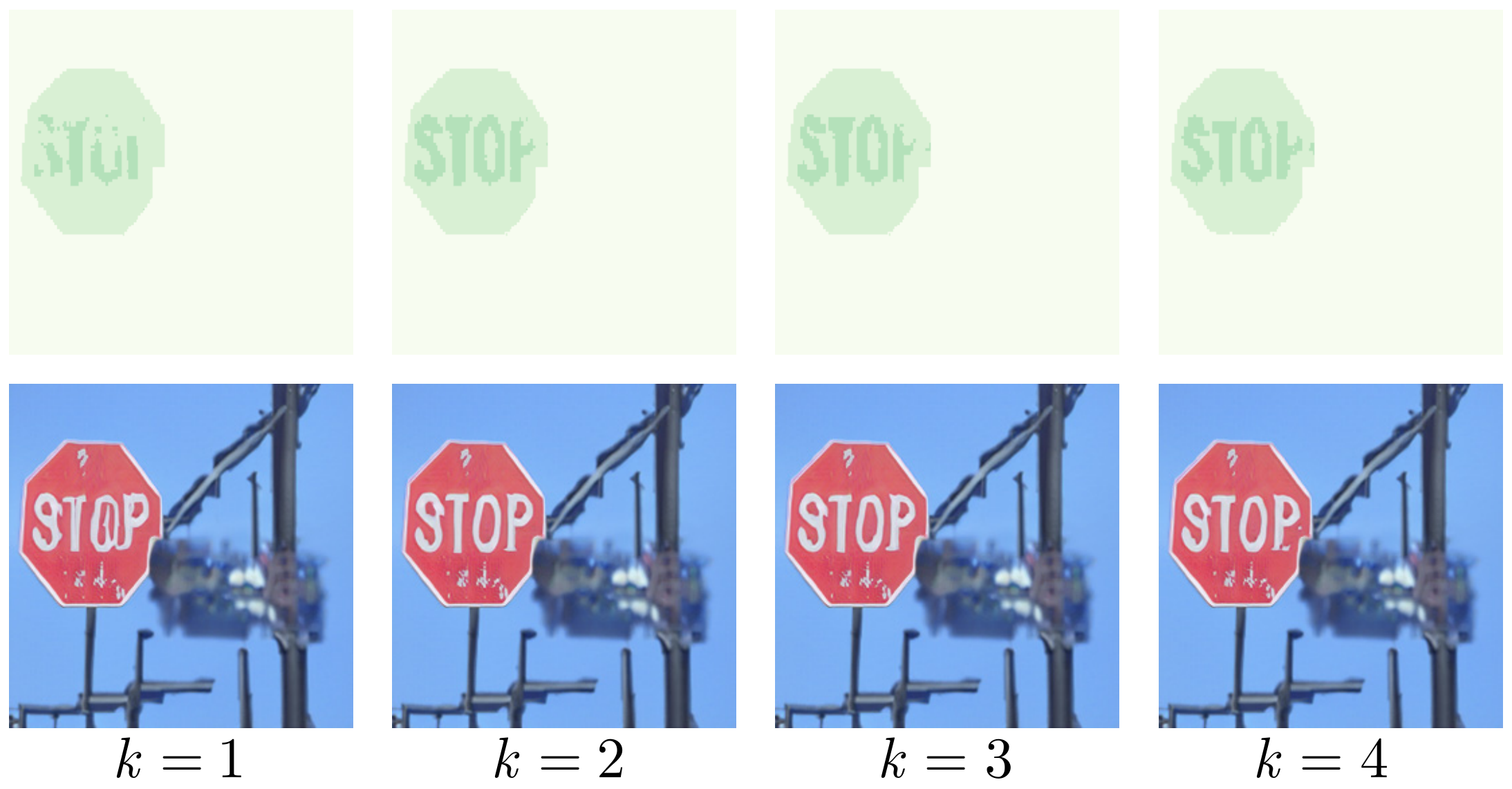}
\vspace{-2em}
\caption{Visual comparisons on the effect of different number of support part maps.}
\vspace{-1.8em}
\label{fig:ablation_shot} 
\end{figure}

\begin{table}[t]
\begin{center}
\caption{Ablation study on the number of support part maps. Basis AC and Novel AC denote the testing accuracy on basis and novel categories of object part dataset, repectively. We also list the corresponding FID score on COCO dataset.}
\vspace{-0.5em}
{
\resizebox{1.0\linewidth}{!}{
    \label{tab:nsupport}
    \setlength{\tabcolsep}{4mm}
    \begin{tabular}{ccccc}
        \toprule
        Methods & 1-shot & 2-shot & 3-shot & 4-shot \\
        \midrule
        Basis AC ($\uparrow$) & 94.30 & 94.37 & 94.38 & 94.37 \\
        Novel AC ($\uparrow$) & 81.41 & 84.48 & 85.00 & 85.16 \\
        COCO FID ($\downarrow$) & 15.76 & 15.73 & 15.72 & 15.72 \\
        \bottomrule
    \end{tabular}
}
}	
\vspace{-2em}
\end{center}
\end{table}

\subsection{Ablation Studies}

We conduct the ablation studies on Cityscapes to verify the effectiveness of our introduced part map and losses.
Furthermore, the effect of the number of support part maps is also analyzed.

\noindent \textbf{Effectiveness of part map and losses.} 
To demonstrate the effectiveness of our proposed part map and losses, we conduct a comparison with the following variants.
(i) Baseline. We adopt the OASIS~\cite{schonfeld2021you} as our baseline. 
(ii) We introduce the global adversarial loss $\mathcal{L}^{g}_{G/D}$ to the baseline model.
(iii) We further introduce the object-level CLIP style loss $\mathcal{L}_{style}$.
(iv) Ours. We finally incorporate the part map and the part semantic modulation resblock to improve image synthesis.
The results are listed in Table~\ref{tab:ablations_results} and Fig.~\ref{fig:ablation_loss}.
From the table, with introduction of $\mathcal{L}^{g}_{G/D}$, $\mathcal{L}_{style}$, and the part map with PSM, the quality of generated images can be improved gradually. 
As shown in Fig.~\ref{fig:ablation_loss}, $\mathcal{L}^{g}_{G/D}$ improves the layout of generated images, and $\mathcal{L}_{style}$ can further enhance the texture synthesis.
Finally, with the introduced part map and PSM Resblock, our method can generate images with high-quality and more photo-realistic parts.
Furthermore, we have cropped and resized each object to $128 \times 128$ to calculate the object-level FID.
From Table~\ref{tab:ablations_results}, our Part\&PSM brings biggest improvement to object-level FID, demonstrating its effectiveness.

\noindent \textbf{Effectiveness of Part Injection.}
We further analyze the effect of the injection method for the part map, and compare it with two variants.
(i) SPADE. We concatenate the part map with semantic map and 3D noise, and use SPADE to modulate the activations.
(ii) Concat. We use part as input, and concatenate it with the input feature of each layer.
From the Table~\ref{tab:ablations_results} and Fig.~\ref{fig:ablation_injection}, injecting the part map through SPADE and Concat is limited in leveraging part map, and the results degenerated to some extent.
In contrast, our methods use the part map and semantic map to modulate the activations sequentially, and achieves better synthesis quality.

\noindent \textbf{Effectiveness of the number of support part maps.} 
We have also conducted the ablations on the number of support part maps.
Specifically, we train the model with mixed numbers of supports ($k=1\sim4$), and test it with different numbers of supports, respectively.
From Table~\ref{tab:nsupport} and Fig.~\ref{fig:ablation_shot}, with the increasing support number, the testing accuracy on both basis and novel categories increases, resulting in more plausible part maps and photo-realistic images.
Similar trends with the FID on COCO dataset.
Moreover, our generator has the ability to fix the incorrect part map to generate plausible results (\eg, $k=1$).
To balance the efficiency and the performance, we select $k = 3$ in default as our final model.
More ablations can be found in the \emph{Suppl}.

% \begin{table}[t]
% \begin{center}
% \caption{Ablation study.}
%     \resizebox{1.0\linewidth}{!}{
%     \small
%     \label{tab:nsupport}
%     \setlength{\tabcolsep}{2.5mm}
%     \begin{tabular}{ccccc}
%         \toprule
%         Methods & 1 shot & 2 shot & 3 shot & 4 shot \\
%         \midrule
%         Closeset AC & 94.30 & 94.37 & 94.38 & 94.37 \\
%         Openset AC & 81.41 & 84.48 & 85.00 & 85.16 \\
%         COCO FID & 15.76 & 15.73 & 15.72 & 15.72 \\
%         \bottomrule
%     \end{tabular}
%     }	
% \end{center}
% \end{table}

\section{Conclusion}

In this paper, we propose a novel method, termed iPOSE, to exploit the part-level information for improving semantic image synthesis. 
A PartNet is first proposed to infer the part map from object shapes based on pre-defined support part maps.
A part semantic modulation Resblock is then presented to leverage the inferred part map with semantic map to perform structure-texture disentangled image synthesis.
With the further introduced global adversarial and object-level CLIP style losses, our iPOSE can generate photo-realistic images, especially for the object parts.
Experiments show that our iPOSE performs favorably against the state-of-the-art methods on the three challenging datasets both qualitatively and quantitatively.

\noindent \textbf{Acknowledgement.} This work was supported in part by National Key R\&D Program of China under Grant No. 2020AAA0104500, and by the National Natural Science Foundation of China (NSFC) under Grant No.s U19A2073 and 62006064.

%%%%%%%%% REFERENCES
{\small
\bibliographystyle{ieee_fullname}
\bibliography{egbib}
}

\clearpage

\section*{A. Object Part Dataset}
\label{sec:dataset}
\subsection*{A.1. Dataset}

\begin{figure}[t]
\centering
\includegraphics[width=1\linewidth]{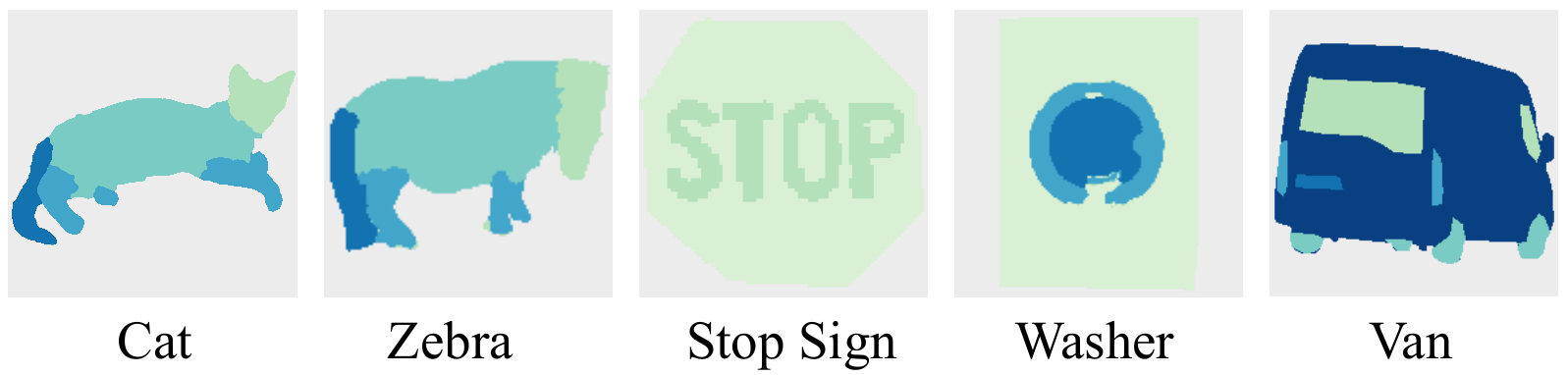}
\vspace{-1.8em}
\caption{Examples of annotated support part maps for semantic image synthesis.}
\label{fig:openset}
\vspace{-1em}
\end{figure}

We build the object part dataset based on the Cityscapes PPS and Pascal VOC PPS datasets proposed by Geus~\etal~\cite{de2021part}.
Specifically, each object part is cropped, resized to 64$\times$64 based on its bounding box, and formed as paired (object shape, object part map). 
For categories in Pascal VOC PPS, some parts are merged as one for simplicity.
For example, we have combined the parts of quadrupeds into four parts (\ie, head, torso, leg, and tail), while combining the parts of cars into five parts (\ie, window, wheel, light, license, and chassis).
The final annotated parts of each category are reported in Table.~\ref{tab:part_labels}.
%
% There is a total of 21 categories in the constructed dataset, and for each category, we form the training/testing set based on its official training/validation partition.
There is a total of 21 categories in the constructed dataset.
For images in Cityscapes PPS, we form the training/testing set based on its official training/validation partition.
For images in Pascal VOC PPS, we merge all images and apportion them into training and testing sets, with an 80-20 split.
%
% Following~\cite{shaban2017one}, we use 20 basis categories for PartNet training (\eg, human, car, bus, and sheep, \etc) and use the remaining category as the novel category for openset validation (\ie, cat).
Following~\cite{shaban2017one}, the total categories are split into basis categories (20 categories, \eg, human, car, bus, and sheep, \etc) and novel category (1 category, \ie, cat).
The training set of basis categories is used to train the PartNet, and the testing set of both basis and novel categories are used to test it.
Besides, for those novel categories in semantic image synthesis datasets, we have annotated $k$ part maps manually as supports to perform part prediction, including the washer, van, zebra, cat, and stop sign.
The total basis/novel/SIS categories are listed in Table~\ref{tab:part_datasets}. 
Fig.~\ref{fig:openset} also illustrates the examples of annotated SIS support part maps.

\begin{table}[t]
\begin{center}
\caption{Category split for object part dataset. We use the training set of basis categories to train the PartNet, and test it on the testing sets of both basis and novel (validation) categories. Novel classes (SIS Testing) denotes the novel classes used for semantic image synthesis.}

\label{tab:part_datasets}
{\small
    \begin{tabular}{lp{5.5cm}}
        \toprule
        Basis Classes & Aeroplane, Bicycle, Bird, Boat, Bottle, Bus, Car, Chair, Cow, Table, Dog, Horse, MotorBike, Human, PottePlant, Sheep, Sofa, Television, Train, Truck \\
        \midrule
        \makecell[c]{Novel Classes \\ (Validation)} & Cat \\
        \midrule
        \makecell[c]{Novel Classes \\ (SIS Testing)} & Washer, Van, Stop sign, Zebra, Cat \\
        \bottomrule
    \end{tabular}
}
\end{center}
\end{table}

\begin{table}[t]
\begin{center}
\caption{Part Annotation Labels}

\label{tab:part_labels}
{\small
    \begin{tabular}{lp{5.5cm}}
        \toprule
        aeroplane & body, stern, wing, wheel \\
        bicycle & wheel, saddle, handlebar, other \\
        bird & head, torso, leg, tail \\
        boat & boat \\
        bottle & cap, body \\
        bus & window, wheel, light, license, chassis \\
        car & window, wheel, light, license, chassis \\ 
        cat & head, torso, leg, tail \\
        chair & chair \\
        cow & head, torso, leg, tail \\
        table & table \\
        dog & head, torso, leg, tail \\
        horse & head, torso, leg, tail \\
        motorbike & wheel, handlebar, saddle, headlight \\
        human & head, torso, leg, arm \\
        pottedplant & pot, plant \\
        sheep & head, torso, leg, tail \\
        train & headlight, torso \\
        sofa & sofa \\
        tvmonitor & screen, frame \\
        truck & window, wheel, light, license, chassis \\ 
        washer & door glass, door, machine body \\
        van & window, wheel, light, license, chassis \\ 
        stop sign & word, octagon \\
        zebra & head, torso, leg, tail \\
        \bottomrule
    \end{tabular}
}
\end{center}
\end{table}

% Basis classes: Aeroplane, Bicycle, Bird, Boat, Bottle, Bus, Car, Chair, Cow, Table, Dog, Horse, MotorBike, HUman, PottePlant, Sheep, Sofa, Television, Truck
% Novel Classes: Cat
% SIS Classes: Washing Machine, Van, Stop sign, zebra, Cat

\subsection*{A.2. Selection of Support Part Maps}

To obtain the support object part maps for each category, we use k-means to cluster the training object shapes into $k$ clusters based on the shape similarity metric~\cite{wang2020constrained}.
To measure the similarity between two object shapes ($O_i$ and $O_j$), we adopt the geometric score~\cite{wang2020constrained} to measure shape consistency,
\begin{equation}
    \sigma(O_i,O_j) = \frac{\Vert O_i - O_j \Vert^2_2}{max\left( \Vert O_i\Vert_1,\Vert O_j \Vert_1) \right)}.
\end{equation}
Lower $\sigma(O_i, O_j)$ indicates more similarity between two object shapes. 
After clustering, part maps with the corresponding object shape closest to the cluster center are selected as support part maps.

\begin{figure}[t]
\centering
\includegraphics[width=1\linewidth]{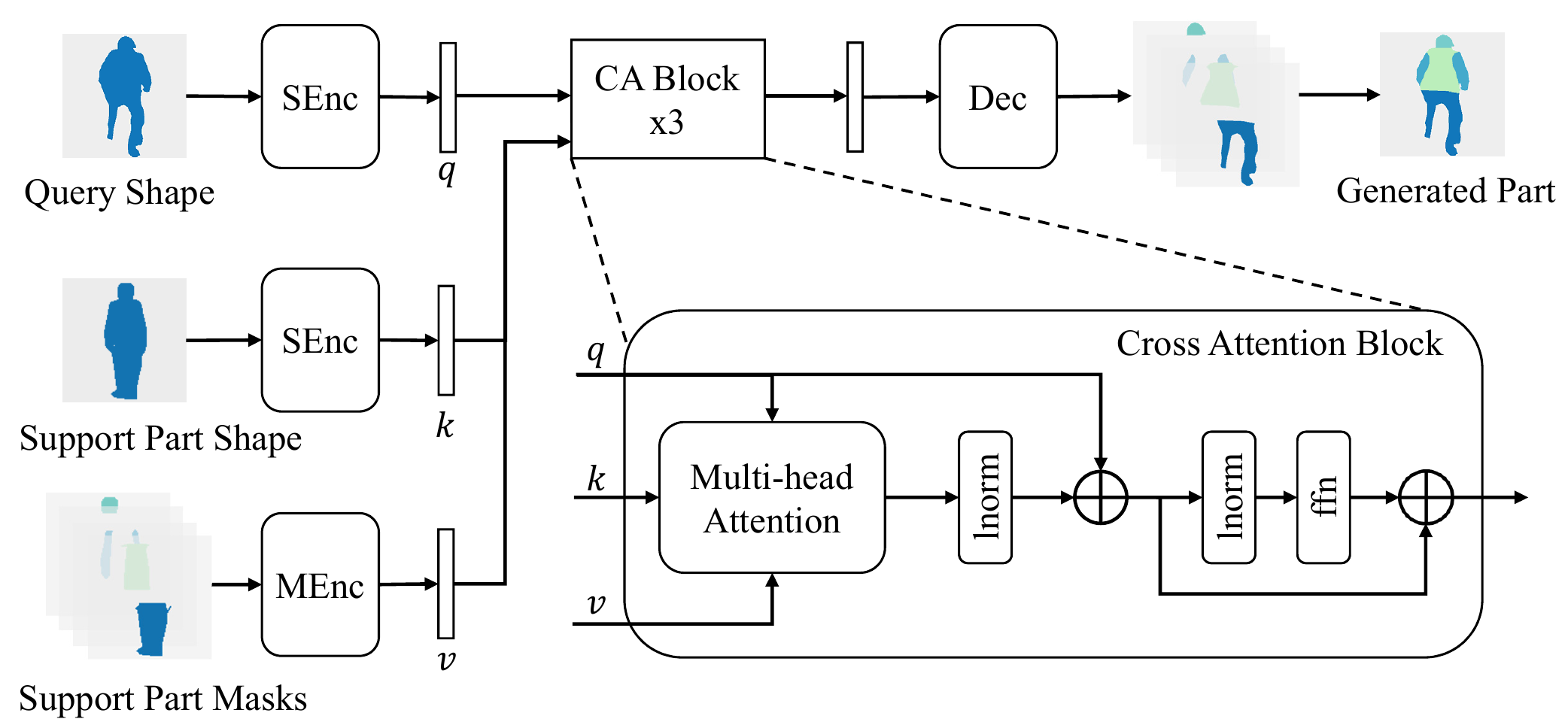}
\caption{Illustration of our PartNet. The support part map is first decomposed into the support part shape and the support part masks as inputs. Cross attention block is adopted to aggregate the part information from the support features.}
\label{fig:partnet1}
\end{figure}

\section*{B. Details of Part Prediction Network}
\label{sec:partnet}

\subsection*{B.1. Network Architecture}

As shown in Fig.~\ref{fig:partnet1}, our PartNet takes the query shape $O_q \in \mathbb{R}^{64\times64\times3}$, support part shape $O^S_{y_q} \in \mathbb{R}^{64\times64\times3}$, and support part masks $O^M_{y_q} \in \mathbb{R}^{64\times64\times1}$ as inputs to predict the part map $P_q \in \mathbb{R}^{64\times64\times1}$.
In particular, it consists of the shape and mask encoders, cross attention blocks, and the decoder, which will be introduced in the following.

\noindent \textbf{Shape Encoder.} 
The shape encoder is composed of 5 Conv-BN-ReLU layers, where Conv denotes the convolutional layer and BN is the batch normalization layer.
For each convolutional layer, the stride is set to 2 to downsample the features.
As mentioned in the main paper, the encoders adopt the multi-scale mechanism to perceive the pixels' relative position of the whole object shape.
Specifically, the outputs of the last two layers are upsampled and concatenated with the output of the third layer as the final output.
Experiments have demonstrated its effectiveness (see Sec.~\ref{sec: multi}).

\noindent \textbf{Mask Encoder.} 
The mask encoder adopts the same architecture as the shape encoder except for the different input channels.

\noindent \textbf{Decoder.} 
The decoder consists of 3 DeConv-BN-ReLU layers, where DeConv denotes the transpose convolutional layer.
For each transpose convolutional layer, the stride is set to 2 to upsample the features.

\subsection*{B.2. Learning Objective}

To facilitate the PartNet learning, we adopt two losses during training.
Firstly, a BCE loss is introduced to encourage the predicted part map to be similar to the ground-truth part map,
\begin{equation}
\mathcal{L}_{pre} =  \text{BCE}(\text{PartNet}(O_q, S_{y_q}), P^{gt}_q),
\end{equation}
where $P^{gt}_q$ denotes the ground-truth part map.
Besides, when the support shape is the same as the query shape, the predict parts should also be the same as the support parts,
\begin{equation}
\mathcal{L}_{rec} =  \text{BCE}(\text{PartNet}(O_q, P^{gt}_q), P^{gt}_q).
\end{equation}
The final learning objective for PartNet is,
\begin{equation}
\mathcal{L}_{part} =  \mathcal{L}_{pre} + \mathcal{L}_{rec}.
\end{equation}

\subsection*{B.3. Experimental Details} 
We train our PartNet on one Tesla V100 GPU and adopt Adam optimizer with $\beta_1$ = 0 and $\beta_2$ = 0.999 where the learning rate is set to 0.0001.
The number of support part maps is set to $k = 3$.
Our PartNet is pre-trained for 30 epochs and fixed during the synthesis training.
Pixel ACcuracy (AC) is adopted as the metric to evaluate the PartNet.

\begin{table}[t]
\begin{center}
\caption{Ablation studies on the multi-scale encoder. With or w/o MS denotes the PartNet with or without the multi-scale encoder. Basis AC and Novel AC denote the testing accuracy on the basis and novel categories of the object part dataset, respectively.}
\vspace{-0.8em}
{\small
    \label{tab:ablation_ms}
    \setlength{\tabcolsep}{4.6mm}
    \begin{tabular}{ccc}
        \toprule
        Object Loss & Basis AC $\left(\uparrow\right)$ & Novel AC $\left(\uparrow\right)$ \\
        \midrule
         w/o MS & 94.05 & 83.21  \\
         with MS & \textbf{94.38} & \textbf{85.00}  \\
        \bottomrule
    \end{tabular}
}
\end{center}
\end{table}

\begin{figure}[t]
\centering
\includegraphics[width=1\linewidth]{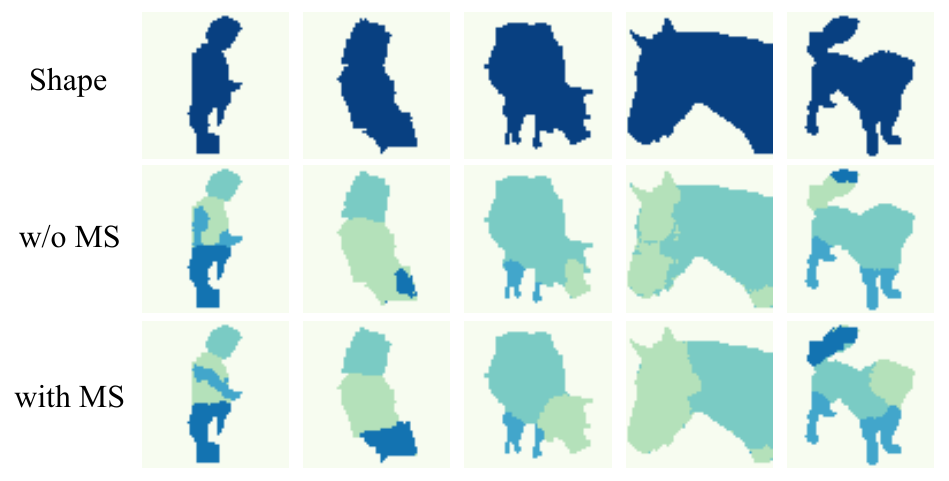}
\caption{Visual comparisons on the effect of the multi-scale encoder. With or w/o MS denotes the PartNet with or without the multi-scale encoder.}
\label{fig:ablation_ms}
\end{figure}

\section*{C. More Ablation Studies on PartNet}
\label{sec:ablation_partnet}

\noindent \textbf{Effectiveness of Multi-Scale Encoder.}
\label{sec: multi}
We first conduct the ablation study on PartNet to verify the effectiveness of the introduced multi-scale encoder.
For comparison, we also train the PartNet without the multi-scale mechanism that the encoders only consist of 3 layers.
As shown in Fig.~\ref{fig:ablation_ms}, without the multi-scale encoder to perceive the whole object shape, the predicted part maps are usually incomplete and discontinuous.
In contrast, with the multi-scale encoder, our PartNet predicts more plausible and realistic part maps, and also achieves better performance on both basis and novel categories (see Table.~\ref{tab:ablation_ms}).

\begin{figure}[t]
\centering
\includegraphics[width=1\linewidth]{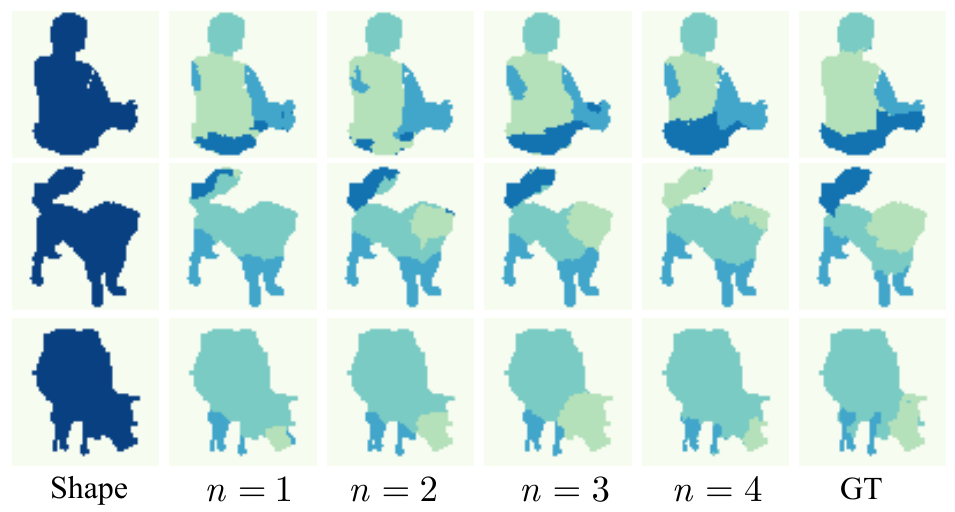}
\caption{Visual comparisons on the effect of different numbers of
cross attention blocks. }
\label{fig:ablation_ncablock}
\end{figure}

\begin{table}[t]
\begin{center}
\caption{Ablation study on the number of cross attention blocks. $n$ denotes the number of blocks.}
{\small
    \label{tab:ncablocks}
    \setlength{\tabcolsep}{3.2mm}
    \begin{tabular}{ccccc}
        \toprule
        Methods & $n=1$ & $n=2$ & $n=3$ & $n=4$ \\
        \midrule
        Basis AC ($\uparrow$) & 94.10 & 94.25 & \textbf{94.38} & 94.34 \\
        Novel AC ($\uparrow$) & 82.28 & 83.76 & \textbf{85.00} & 83.41 \\
        \bottomrule
    \end{tabular}
}	
\end{center}
\end{table}

\noindent \textbf{Effectiveness of the number of CA blocks.}
Furthermore, the effect of the number of cross attention (CA) blocks is also analyzed.
We train the PartNet with the different number of CA blocks, and the results are listed in Fig.~\ref{fig:ablation_ncablock} and Table~\ref{tab:ncablocks}.
From Table~\ref{tab:ncablocks}, more cross attention blocks bring more prediction capabilities to the PartNet, resulting in better prediction accuracy on both basis and novel categories. 
However, when the number $n > 3$, the PartNet tends to be overfitting, and adding more cross attention blocks will not bring more performance gain.
Thus, we choose PartNet with three cross attention blocks as our final part prediction model.

\begin{figure*}[t]
\centering
\includegraphics[width=1\linewidth]{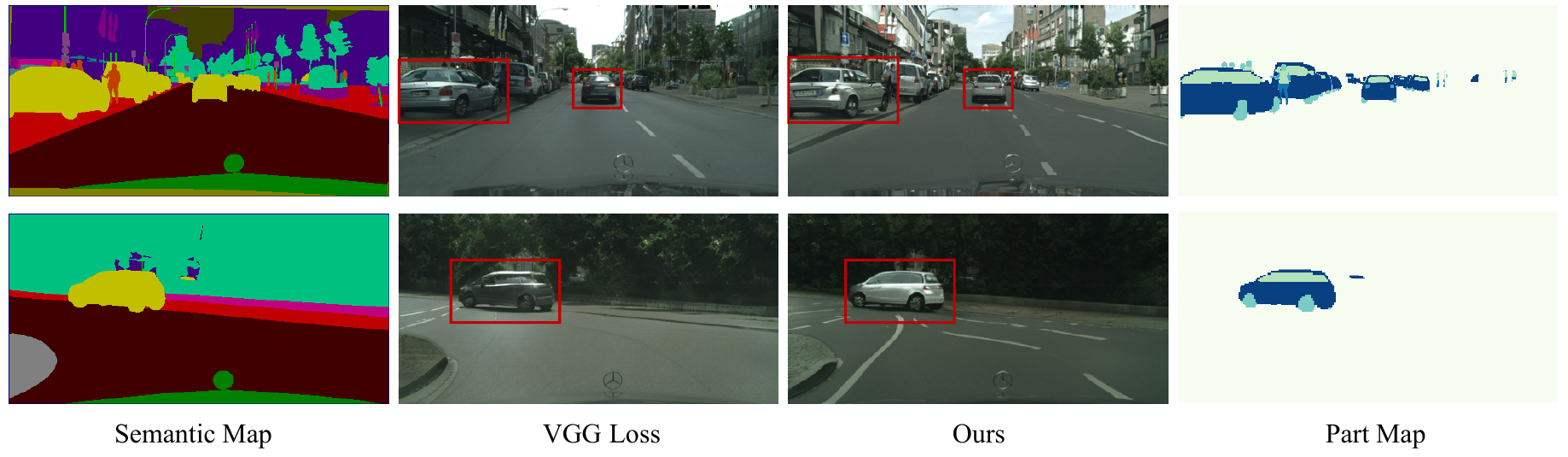}
\caption{Visual comparisons between CLIP style loss and VGG loss.}
\label{fig:ablation_clip}
\end{figure*}

\begin{figure*}[t]
\centering
\includegraphics[width=1\linewidth]{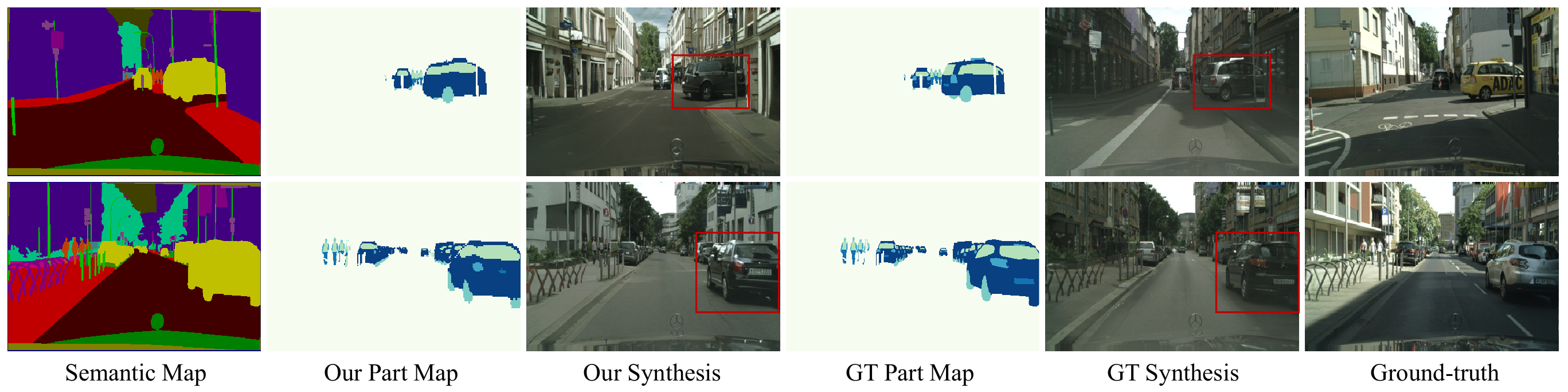}
\caption{Visual comparisons between the images generated with predicted part maps and the images generated with ground-truth part maps.}
\label{fig:ablation_partmap}
\end{figure*}

\section*{D. More Ablation Studies on Synthesis}
\label{sec:ablation_synthesis}

\noindent \textbf{CLIP Style Loss vs. VGG Loss.}
To demonstrate the effectiveness of the object-level CLIP style loss~\cite{zhang2022towards}, we further compare it with an object-level VGG loss~\cite{wang2018high}.
Specifically, for CLIP style loss, we adopt the pre-trained CLIP image encoder (VIT-32)~\cite{radford2021learning} as the feature extractor, and the tokens of the eighth layer are used to calculate the loss.
Each object of the generated images is cropped and resized to $224 \times 224$ as input.
For VGG loss, we adopt the pre-trained VGG19~\cite{simonyan2014very} as the feature extractor, and the intermediate features are used to calculate the loss.
Each object of the generated images is cropped and resized to $128 \times 128$ as input.
The results are listed in Fig.~\ref{fig:ablation_clip} and Table~\ref{tab:ablation_clip}.
As shown in Fig.~\ref{fig:ablation_clip}, benefited from the large-scale pre-training of CLIP, our iPOSE trained with CLIP style loss achieves better visual quality.
%
% our iPOSE with CLIP style loss
%
Besides, it has the ability to refine the part map to generate images with more realistic parts.
From Table~\ref{tab:ablation_clip}, our iPOSE with CLIP style loss also performs better than VGG loss, demonstrating its effectiveness.

% 41.58/68.57/81.72

\begin{table}[t]
\begin{center}
\caption{Ablation study on the object-level CLIP style loss, and we further compare it with an object-level VGG loss.}
{\small
    \label{tab:ablation_clip}
    \setlength{\tabcolsep}{4.6mm}
    \begin{tabular}{cccc}
        \toprule
        Object Loss & FID$\left(\downarrow\right)$ & mIOU$\left(\uparrow\right)$ & AC$\left(\uparrow\right)$  \\
        \midrule
         VGG Loss & 41.6 & 68.6 & 81.7 \\
         CLIP Loss & \textbf{41.3} & \textbf{70.6} & \textbf{82.2} \\
        \bottomrule
    \end{tabular}
}
\end{center}
\end{table}

\begin{table}[t]
\begin{center}
\caption{Ablation studies of losses and part map on Cityscapes.}
\resizebox{1\linewidth}{!}{
    \label{tab:ablations_results_more}
    \setlength{\tabcolsep}{2.5mm}
    \begin{tabular}{ccccccc}
        \toprule
        Part \& PSM & $\mathcal{L}^g_{G/D}$ & $\mathcal{L}_{style}$ & FID$\left(\downarrow\right)$ & mIOU$\left(\uparrow\right)$ & AC$\left(\uparrow\right)$ & obj FID $\left(\downarrow\right)$  \\
        \midrule
        &  &  & 47.7 & 66.9 & 81.5 & 44.1 \\
        \midrule
        \checkmark &  &  & 44.2 & 69.0 & 81.8 & 36.8 \\
        & \checkmark &  & 43.6 & 66.7 & 81.9 & 39.2 \\
        &  & \checkmark & 45.2 & 69.1 & 81.2 & 38.9 \\ 
        & \checkmark & \checkmark & 42.8 & 70.5 & 82.1 & 37.5 \\
        \checkmark &  & \checkmark & 43.1 & \textbf{70.6} & 82.0 & 35.5  \\
        \midrule
        \checkmark & \checkmark & \checkmark & \textbf{41.3} & \textbf{70.6} & \textbf{82.2} &  \textbf{30.4} \\
        \bottomrule
    \end{tabular}
}	
\end{center}
\end{table}

\begin{table}[t]
\begin{center}
\caption{Comparison of our part map and SPD feature~\cite{lv2022semantic}. }
\label{tab:spd}
\resizebox{1\linewidth}{!}{
    \renewcommand{\arraystretch}{1.2}
    \begin{tabular}{lccc | lccc}
        \toprule
        Method & FID$\left(\downarrow\right)$ & mIOU$\left(\uparrow\right)$ & AC$\left(\uparrow\right)$ & Method & FID$\left(\downarrow\right)$ & mIOU$\left(\uparrow\right)$ & AC$\left(\uparrow\right)$  \\
        \midrule
        Ours & \textbf{41.2} & \textbf{70.6} & \textbf{82.2} & Ours w/SPD & 43.1 & 70.3 & 82.0  \\
        \bottomrule
    \end{tabular}
}
\end{center}
\end{table}

\begin{figure}[t]
\centering
\includegraphics[width=1\linewidth]{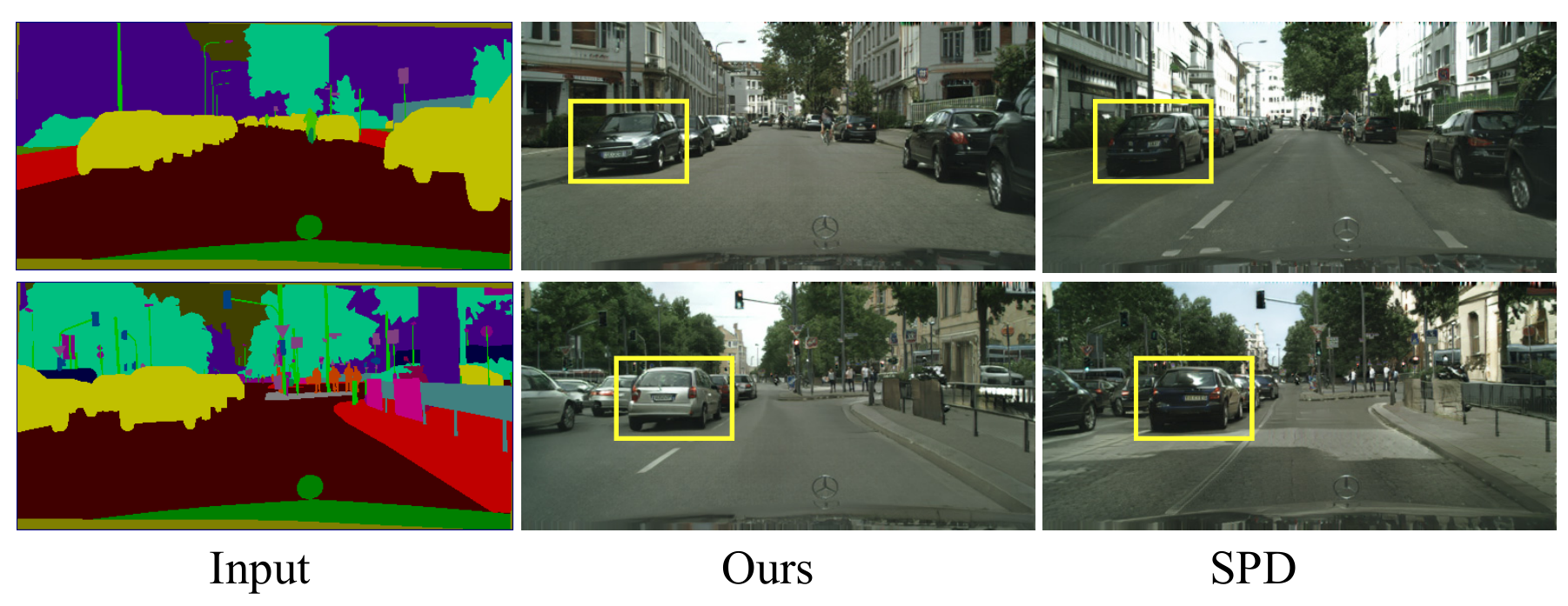}
\caption{Visual comparisons between the images generated with predicted part maps and the images generated with SPD~\cite{lv2022semantic}.}
\label{fig:ablation_spd}
\end{figure}

\noindent \textbf{Predicted Part Map vs. SPD.}
We have conducted the experiment by replacing our part map with SPD feature~\cite{lv2022semantic} on Cityscapes, while keep same network architecture. 
From Fig.~\ref{fig:ablation_spd}(a) and Table~\ref{tab:spd}, our iPOSE generates more realistic parts, and also performs favorably against SPD, especially on FID.

\noindent \textbf{Predicted Part Map vs. GT Part Map.}
We also compare the predicted part maps with the ground-truth part maps.
Since there are part annotations on Cityscapes~\cite{de2021part}, we additionally train a model with ground-truth part maps as inputs for comparison.
% %
The results are listed in Fig.~\ref{fig:ablation_partmap} and Table~\ref{tab:ablation_gt}.
As shown in Fig.~\ref{fig:ablation_partmap}, with the ground-truth part map as guidance, the model generates images with accuracy parts.
While our iPOSE with the predicted part map can also generate images with realistic parts, even for those objects with extreme poses (the first row in Fig.~\ref{fig:ablation_partmap}).
From Table~\ref{tab:ablation_gt}, our iPOSE achieves comparable performance compared to the ground-truth part map as input.

\begin{table}[t]
\begin{center}
\caption{Quantitative comparisons between the results generated by the predicted part maps and the results generated by the ground-truth part maps.}
{\small
    \label{tab:ablation_gt}
    \setlength{\tabcolsep}{4.6mm}
    \begin{tabular}{cccc}
        \toprule
        Part Map & FID$\left(\downarrow\right)$ & mIOU$\left(\uparrow\right)$ & AC$\left(\uparrow\right)$  \\
        \midrule
         Ours  & 41.3 & 70.6 & 82.2 \\
         GT & \textbf{40.8} & \textbf{70.9} & \textbf{82.3} \\
        \bottomrule
    \end{tabular}
}
\end{center}
\end{table}

\begin{table}[t]
\begin{center}
\caption{Object-level user study on different datasets. The numbers indicate the percentage (\%) of volunteers who favor the results of our method over those of the competing methods. }
\vspace{-0.8em}
\label{tab:user_study_obj}
\resizebox{1\linewidth}{!}{
    \setlength{\tabcolsep}{1.2em}
    \begin{tabular}{lcccc}
        \toprule
        Dataset & \makecell[c]{Ours vs. \\ SPADE} & \makecell[c]{Ours vs. \\ CC-FPSE}  & \makecell[c]{Ours vs. \\ OASIS} & \makecell[c]{Ours vs. \\ SAFM} \\
        \midrule
        Cityscapes~\cite{cordts2016cityscapes} & 85.3 & 81.1 & 80.1 & 79.2 \\
        ADE20K~\cite{zhou2017scene} & 81.0 & 74.5 & 71.7 & 60.7 \\
        COCO-Stuff~\cite{caesar2018coco} & 75.2 & 63.1 & 66.5 & 72.0 \\
        \bottomrule
    \end{tabular}
}
\vspace{-1.2em}
\end{center}
\end{table}

\noindent \textbf{More ablations on losses and Part\&PSM.} 
We have conducted the experiments by adding Part\&PSM, $\mathcal{L}_{style}$, and $\mathcal{L}^g_{G/D}$ to baseline respectively, and the results are reported in Table~\ref{tab:ablations_results_more} (first 4 rows).
Among them, our Part\&PSM contributes most to the object synthesis and bring a significant performance improvement, especially on object-level FID.
$\mathcal{L}^g_{G/D}$ brings more improvement on FID, because it improves not only objects, but also the background.
We have further conducted the experiments by adding Part\&PSM, $\mathcal{L}_{style}$, and $\mathcal{L}^g_{G/D}$ sequentially.
From Table~\ref{tab:ablations_results} (rows 1, 2, 6, and 7), our Part\&PSM enables to generate photo-realistic object parts and obtains a significant improvement on FID, mIOU, and obj FID.
$\mathcal{L}_{style}$ and $\mathcal{L}^g_{G/D}$ can further boost the performance.

\section*{E. More Qualitative Results}
\label{sec:more_qualitative_results}

\begin{figure*}[t]
\centering
\includegraphics[width=1\linewidth]{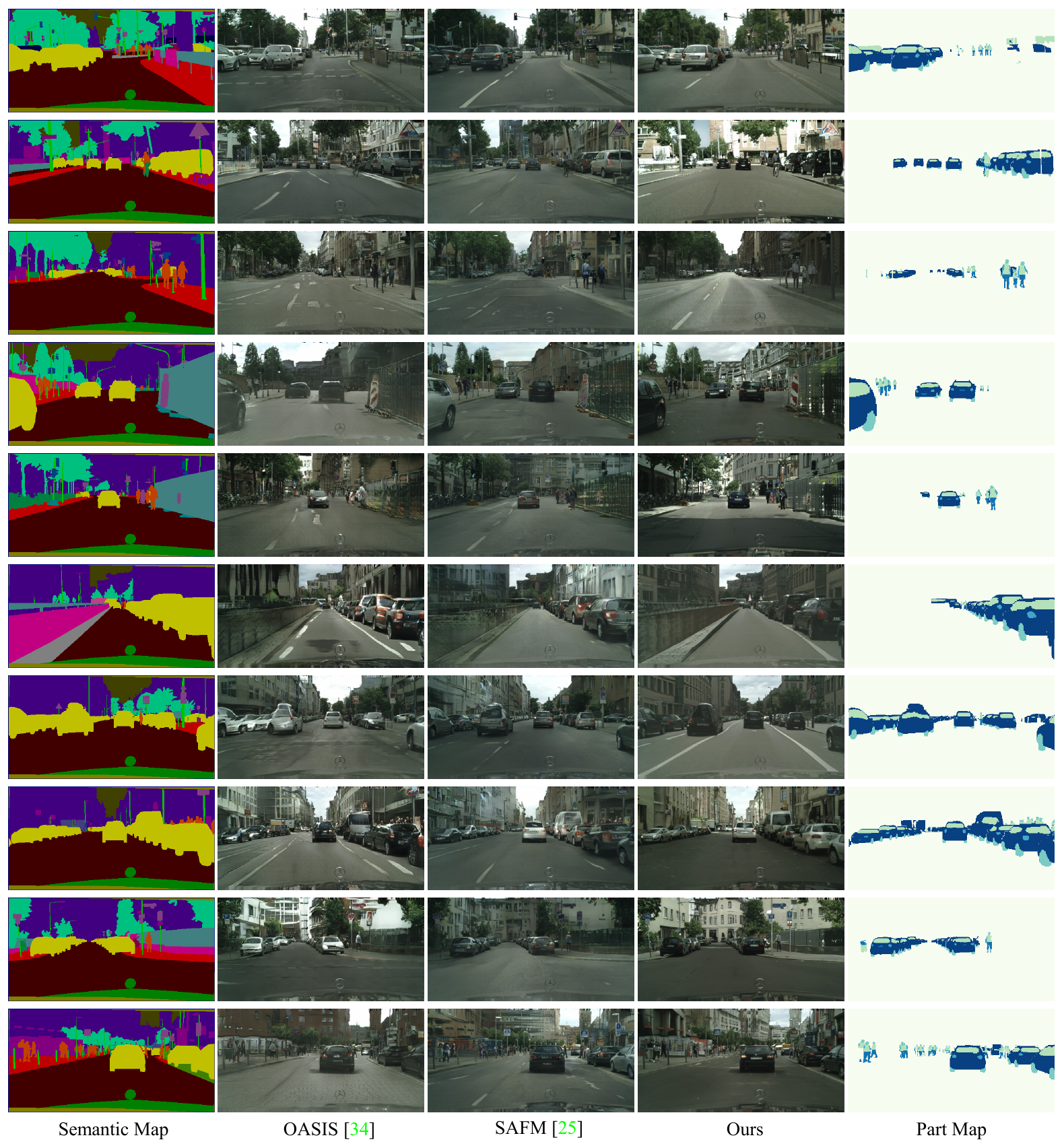}
\vspace{-1.5em}
\caption{More qualitative comparison results on Cityscapes~\cite{cordts2016cityscapes}.}
\vspace{-1.5em}
\label{fig:city_more}
\end{figure*}

\begin{figure*}[t]
\centering
\includegraphics[width=1\linewidth]{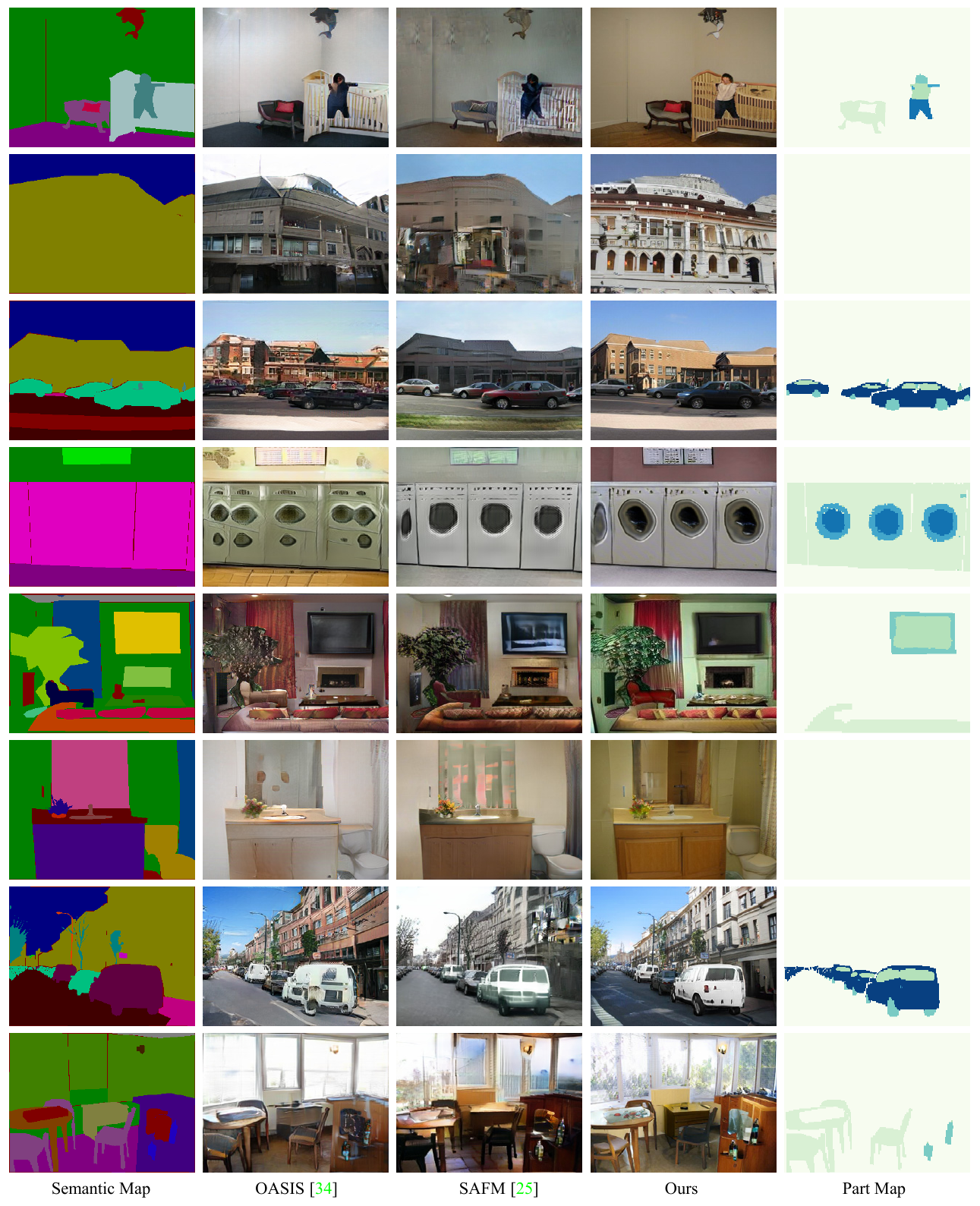}
\vspace{-1.5em}
\caption{More qualitative comparison results on ADE20K~\cite{zhou2017scene}.}
\vspace{-1.5em}
\label{fig:ade_more}
\end{figure*}

\begin{figure*}[t]
\centering
\includegraphics[width=1\linewidth]{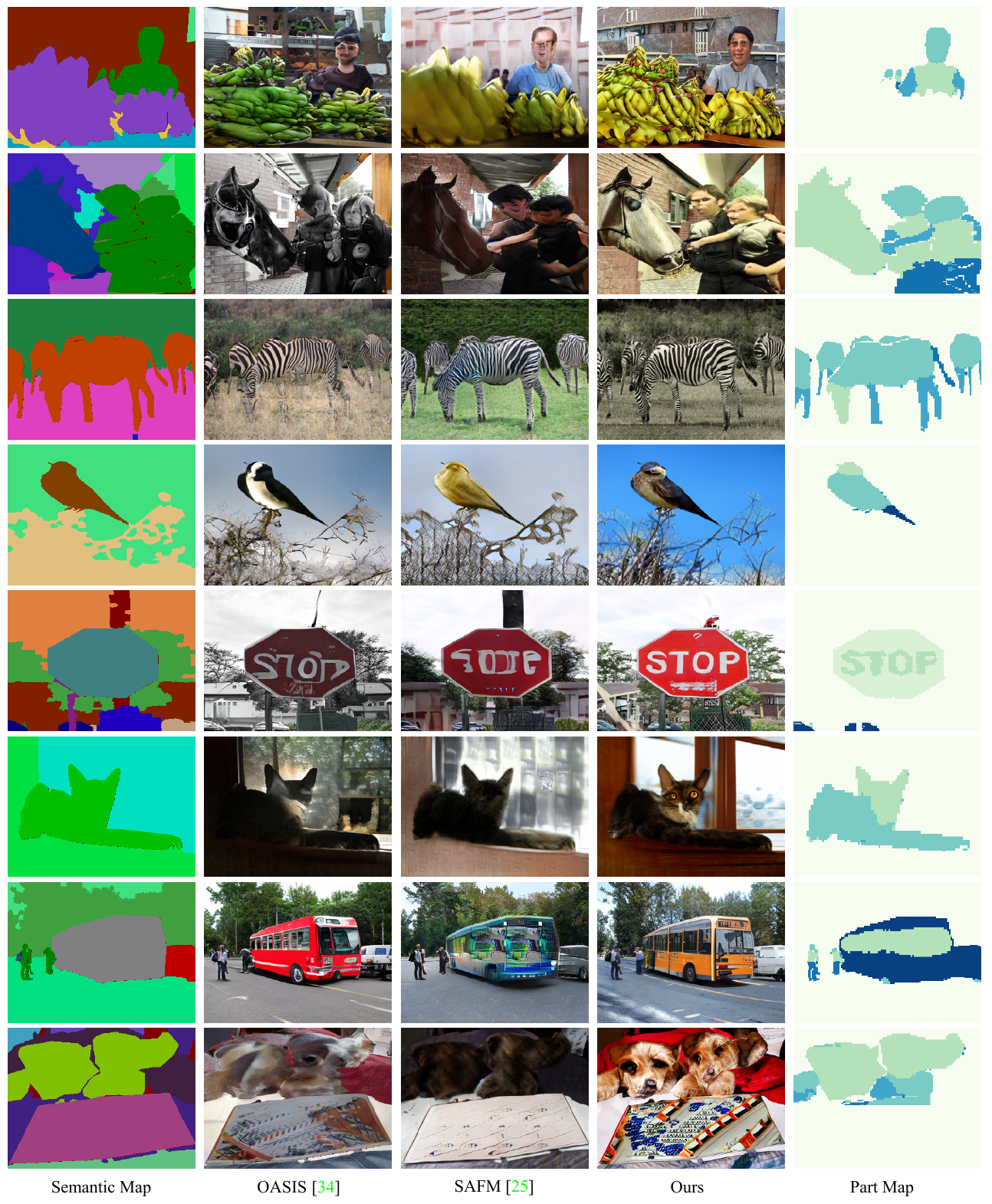}
\vspace{-1.5em}
\caption{More qualitative comparison results on COCO~\cite{caesar2018coco}.}
\vspace{-1.5em}
\label{fig:coco_more}
\end{figure*}

\begin{figure*}[t]
\centering
\includegraphics[width=1\linewidth]{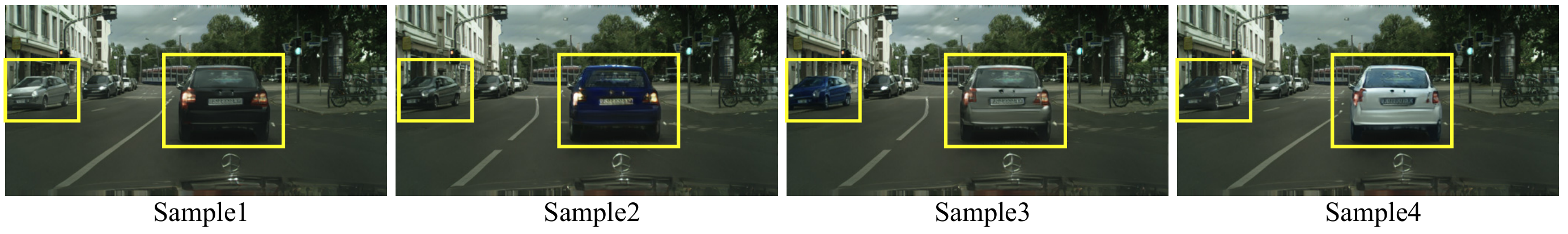}
\vspace{-1.5em}
\caption{Object-level diversity. By sampling different noises for each object, our method can synthesize diverse object results.}
\vspace{-1.5em}
\label{fig:car_diversity}
\end{figure*}

Fig.~\ref{fig:city_more}, Fig.~\ref{fig:ade_more} and Fig.~\ref{fig:coco_more} illustrate the qualitative comparisons between our iPOSE and state-of-the-art methods~\cite{schonfeld2021you, lv2022semantic}. 
As shown in the figures, our iPOSE generates images with more realistic parts, further demonstrating its superiority.
Moreover, as shown in Fig.~\ref{fig:car_diversity}, by sampling different noises for each object, our method can synthesize diverse object results.

\section*{F. More Quantitative Results}
\label{sec:more_quantitative_results}

We have also conducted the object-level user study to evaluate the effects of our iPOSE on object synthesis.
Specifically, we cropped and resized each object to $128 \times 128$ to perform the object-level human evaluation.
From Table~\ref{tab:user_study_obj}, users tend to favor our results on all the datasets.
Besides, compared to the global-level evaluation (Table~{\color{red}2} in main paper), our iPOSE obtains a better preference in object-level.

\end{document}